\documentclass[10pt,twocolumn,letterpaper]{article}

\usepackage{wacv}
\usepackage{times}
\usepackage{epsfig}
\usepackage{graphicx}
\usepackage{amsmath}
\usepackage{amssymb}

\usepackage{microtype}
\usepackage{graphicx}
\usepackage{booktabs}
\usepackage{nicefrac}
\usepackage{amsmath,amsfonts,amssymb,mathrsfs}
\usepackage{bm}
\usepackage{grffile}
\usepackage{xspace}
\usepackage{etoolbox}
\usepackage[T1]{fontenc}

\newcommand{\mathbold}[1]{\bm{#1}}
\newcommand{\mbf}[1]{\mathbf{#1}}
\newcommand{\vect}[1]{\mathbf{#1}}

\newcommand{\T}{^\mathsf{T}}

\newcommand{\vtheta}[0]{\mathbold{\theta}}

\newcommand{\vphi}[0]{\mathbold{\phi}}

\newcommand{\vx}{\mbf{x}}

\newcommand{\vz}{\mbf{z}}

\newcommand{\MI}{\mbf{I}}

\newcommand{\CelebA}{\textsc{CelebA}\xspace}
\newcommand{\CelebAHQ}{\textsc{CelebA-HQ}\xspace}

\newcommand{\LSUN}{\textsc{LSUN}\xspace}
\newcommand{\PIONEER}{\textsc{Pioneer}\xspace}

\graphicspath{ {img/} }

\usepackage[bookmarks=false,pagebackref=false,breaklinks=true,letterpaper=true,colorlinks=true]{hyperref}
  \newcommand{\todo}[1]{\textcolor{blue}{\textbf{[#1]}}}

  \usepackage{xcolor,colortbl}

  \usepackage{booktabs}
  \usepackage{multirow}
  \usepackage{tabularx}

  \makeatletter
  \let\MYcaption\@makecaption
  \makeatother

  \usepackage{subcaption}

  \makeatletter
  \let\@makecaption\MYcaption
  \makeatother

  \usepackage{tikz,pgfplots}
  \usetikzlibrary{plotmarks,shapes,shapes.arrows,positioning}
  \pgfplotsset{compat=newest} 
  \pgfkeys{/pgf/number format/.cd,1000 sep={}}

  \newlength\figureheight
  \newlength\figurewidth

  \pgfplotsset{every axis/.append style={
    grid style={line width=0.6pt,dotted,gray}}}

  \newcommand{\citep}[1]{\todo{#1}}
  \newcommand{\citet}[1]{\todo{#1}}

\wacvfinalcopy

\def\wacvPaperID{684}

\ifwacvfinal\fi
\begin{document}

\title{Towards Photographic Image Manipulation with Balanced \\ Growing of Generative Autoencoders}

\author{Ari Heljakka$^{1,2}$ \qquad Arno Solin$^1$ \qquad Juho Kannala$^1$\\
$^1$Department of Computer Science, Aalto University, Espoo, Finland \\
$^2$GenMind Ltd., Espoo, Finland \\ 
{\tt\small firstname.lastname@aalto.fi}}

\maketitle
\ifwacvfinal\thispagestyle{empty}\fi

\begin{abstract}

  We present a generative autoencoder that provides fast encoding, faithful reconstructions (\eg retaining the identity of a face), sharp generated/reconstructed samples in high resolutions, and a well-structured latent space that supports semantic manipulation of the inputs. There are no current autoencoder or GAN models that satisfactorily achieve all of these. We build on the progressively growing autoencoder model \PIONEER, for which we completely alter the training dynamics based on a careful analysis of recently introduced normalization schemes. We show significantly improved visual and quantitative results for face identity conservation in \CelebAHQ. Our model achieves state-of-the-art disentanglement of latent space, both quantitatively and via realistic image attribute manipulations. On the \LSUN Bedrooms dataset, we improve the disentanglement performance of the vanilla \PIONEER, despite having a simpler model. Overall, our results indicate that the \PIONEER networks provide a way towards photorealistic face manipulation.
  
\end{abstract}

\begin{figure}[!t]
  \centering\scriptsize
  \newlength{\hexsize}
  \setlength{\hexsize}{.9cm}
  \setlength{\figurewidth}{1.8\hexsize}
  \setlength{\figureheight}{\figurewidth}
  \begin{tikzpicture}[inner sep=0]
  \tikzstyle{hexagon} = [rounded corners=3pt,regular polygon,regular polygon sides=6,shape border rotate=90,draw=white,fill=gray,minimum width=2\hexsize]

  \newcommand{\hexagon}[4]{\node (#1) [hexagon,path picture={\node at (path picture bounding box.center){\includegraphics[width=\figurewidth]{fig/fig1/celebaHQ/hex13/hex_interpolations_6_25480000_1.0x#1.jpg}};}] at (#2\hexsize,#3\hexsize) {};};

  \hexagon{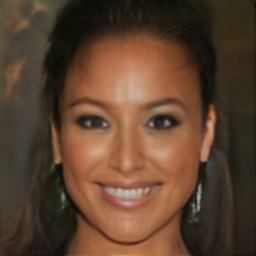}{-1.7321}{-3.0000}{};
  \hexagon{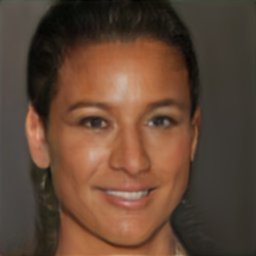}{0.0000}{-3.0000}{};
  \hexagon{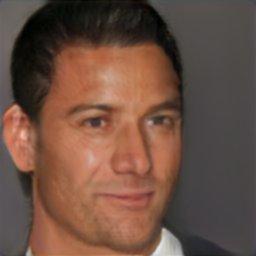}{1.7321}{-3.0000}{};
  \hexagon{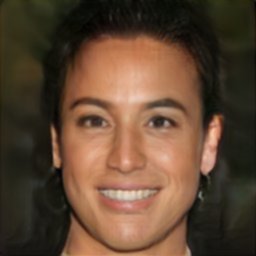}{-2.5981}{-1.5000}{};
  \hexagon{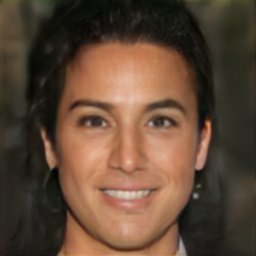}{-0.8660}{-1.5000}{};
  \hexagon{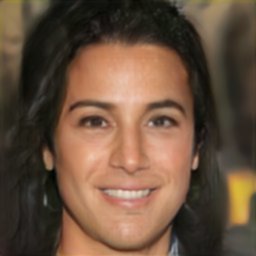}{0.8660}{-1.5000}{};
  \hexagon{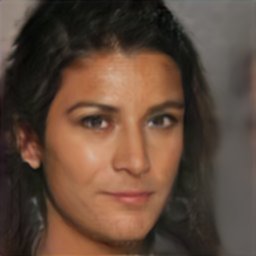}{2.5981}{-1.5000}{};
  \hexagon{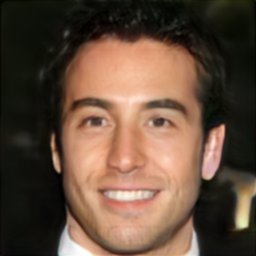}{-3.4641}{0.0000}{};
  \hexagon{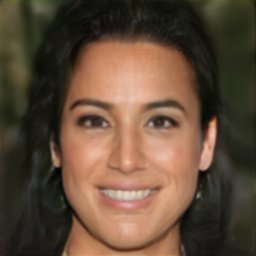}{-1.7321}{0.0000}{};
  \hexagon{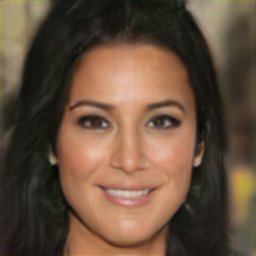}{0.0000}{0.0000}{};
  \hexagon{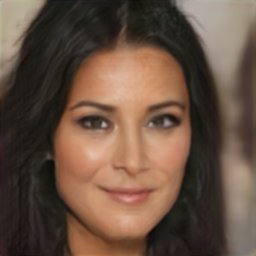}{1.7321}{0.0000}{};
  \hexagon{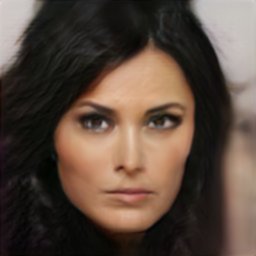}{3.4641}{0.0000}{};
  \hexagon{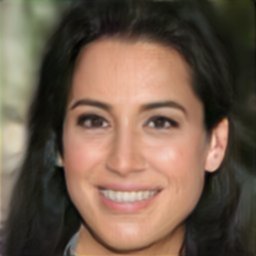}{-2.5981}{1.5000}{};
  \hexagon{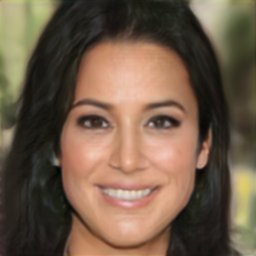}{-0.8660}{1.5000}{};
  \hexagon{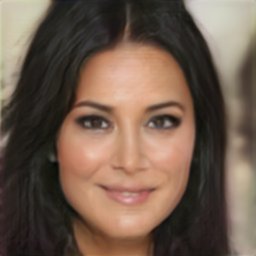}{0.8660}{1.5000}{};
  \hexagon{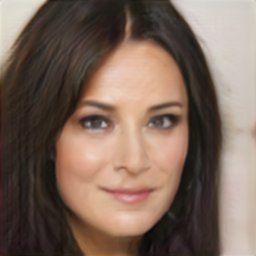}{2.5981}{1.5000}{};
  \hexagon{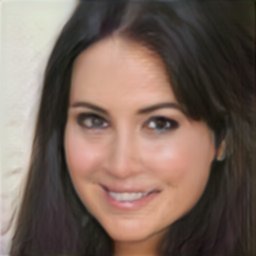}{-1.7321}{3.0000}{};
  \hexagon{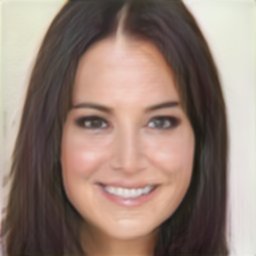}{0.0000}{3.0000}{};
  \hexagon{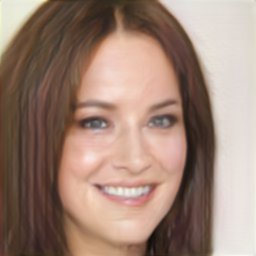}{1.7321}{3.0000}{};

  \newcommand{\insquare}[4]{\node [minimum width=.75*\figurewidth,minimum height=.75*\figureheight, rounded corners=3pt,path picture={\node at (path picture bounding box.center){\includegraphics[width=.75\figurewidth]{fig/fig1/celebaHQ/hex13/hex_interpolations_6_25480000_1.0_orig_#3.jpg}};}] at (#1,#2) {};};
  
  \foreach \x [count=\i] in {4,2,0,1,3,5} {
    \insquare{{-3.5*.8*\figurewidth+\i*.8*\figurewidth}}{-4.6}{\x}{\x}}

  \foreach \x [count=\i] in {a,b,c,d,e,f} {    
    \node at ({-3.5*.8*\figurewidth+\i*.8*\figurewidth},-5.5) {(\x)};}

  \node [left of=16] {(a)};
  \node [left of=7] {(b)};
  \node [left of=0] {(c)};
  \node [right of=2] {(d)};
  \node [right of=11] {(e)};
  \node [right of=18] {(f)};

  \end{tikzpicture}
  \caption{Six-way example interpolations in the latent space between reconstructions of previously unseen input images (bottom) at $256{\times}256$ resolution. Best viewed zoomed in.}
  \label{fig:hex}  
\end{figure}

\begin{figure*}[!t]
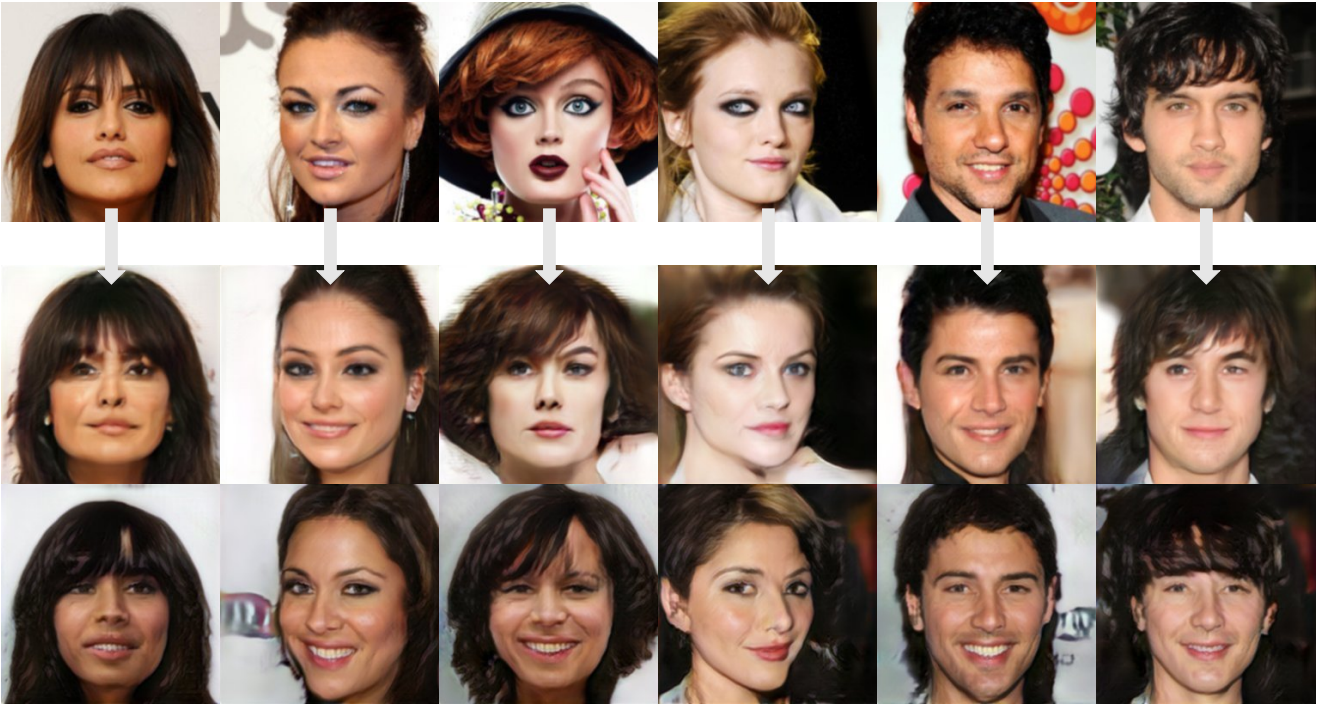

  \centering\footnotesize
  \setlength{\figurewidth}{.1667\textwidth}
  \setlength{\figureheight}{\figurewidth}

  \begin{tikzpicture}[inner sep=0]

  \tikzstyle{arrow} = [draw=black!10, single arrow, minimum height=10mm, minimum width=3mm, single arrow head extend=1mm, fill=black!10, anchor=center, rotate=-90, inner sep=2pt]

  \foreach \x [count=\i] in {9792,9793,9794,9795,9783,120} {
     \node[] at ({\figurewidth*\i},{0*\figureheight}) {\includegraphics[width=\figurewidth]{./fig/fig2/celeba/final/\x_orig.jpg}};
     \node[] at ({\figurewidth*\i},{-1.2\figureheight}) {\includegraphics[width=\figurewidth]{./fig/fig2/celeba/final/\x_pine.jpg}};
     \node[] at ({\figurewidth*\i},{-2.2\figureheight}) {\includegraphics[width=\figurewidth]{./fig/fig2/celeba/final/\x_pine_old.jpg}};
     \node[arrow] at ({\figurewidth*\i},{-0.6\figureheight}) {};}

  \end{tikzpicture}
  \caption{Example of reconstruction quality in $256{\times}256$ resolution with typical images from the \CelebAHQ test set (top row), by our balanced \PIONEER (middle) and baseline \PIONEER (bottom). Here, the input images are encoded into 512-dimensional latent feature vector and decoded back to the original dimensionality (middle and bottom rows). The encoding--decoding of balanced \PIONEER tends to preserve facial features, orientation, expressions, and hair style. Small mistakes can still be observed, especially in male subjects. (Features highly underrepresented in the limited \CelebAHQ training set, such as a dark-skinned face with atypical orientation, will result in more pronounced failures.)}
  \label{fig:reconstructions}
\end{figure*}

\section{Introduction}
\label{sec:intro}
\noindent
Advances in generative image modelling with deep neural networks have raised expectations for delivering new tools for photographic image manipulation and exploration. Generative image modelling involves training the model by a wide variety of data, such as pictures of faces or bedrooms, and allowing it to learn features and structure present in the data. Recent works (\eg \cite{karras2017,karras2018style,brock2018}) on using Generative Adversarial Networks (GANs, \cite{goodfellow2014}) for images of high quality and resolution have showed that {\em generating} random beautiful and sharp high-resolution images is achievable by current tools---given a fair deal of engineering and compute. 

However, to manipulate existing images and other content, we also need inference, not only the capability to generate random samples. A model that can encode the sample of interest into a well-structured latent space allows us to manipulate the sample via that latent representation. In absence of extensions, a GAN has no inference component. In contrast, generative autoencoder models allow both generation and reconstruction, with a bi-directional mapping between the latent feature space and image space. The latent space can be exploited by interpolating between two or more points that represent images (see Fig.~\ref{fig:hex}), by finding directions that encode an individual feature of interest (see Fig.~\ref{fig:manipulate}) to modify new images, and for other downstream tasks.

The state-of-the-art in this direction was set by three papers simultaneously published in late 2018: the flow-based GLOW method \cite{Kingma+Dhariwal:2018}, the introspective variational autoencoder (IntroVAE, \cite{huang2018}), and the progressively growing generative autoencoder (\PIONEER, \cite{heljakka2018}) based on the Adversarial Generator--Encoder (AGE, \cite{ulyanov2017}). GLOW offers tractable likelihood estimates and a degree of attribute manipulation but ultimately suffers from mode collapse issues. IntroVAE offers high sample quality but with \eg human faces, it often only retains overall {\it topology}, not the identity \cite{huang2018}.

Here, we build on the progressively growing generative autoencoder concept, primarily the \PIONEER. We seek to balance the training in higher resolutions---a common challenge for generative models (see \eg discussion in \cite{huang2018}). \PIONEER was previously shown to reconstruct images only up to $128{\times}128$. Our model provides $256{\times}256$ reconstructions (Fig.~\ref{fig:reconstructions}), generates realistic and diverse samples, and learns disentangled latent representations of features (Fig.~\ref{fig:manipulate}).

The contributions and results of this paper are as follows.
{\it (i)}~We show that an AGE-based autoencoder model (balanced \PIONEER) can learn a high-resolution image dataset, combining all the strengths of both autoencoder and GAN models---fast encoding, faithful reconstruction of inputs, and sharp sample generation (contrary to what is implied in \cite{huang2018}). We also show state-of-the-art disentanglement capabilities in terms of Perceptual Path Length metric \cite{karras2018style}, along with image attribute modifications at higher resolutions than previously shown for unsupervised general-purpose models.

{\it (ii)}~We propose a modified \PIONEER model with completely altered training dynamics, considerably improving the learning capacity. When reconstructing face images of \CelebAHQ, the baseline \PIONEER often loses the identity. The cause is related to a large dynamic range and occasional collapse of the encoder--decoder minimax game, complicated by three different weight normalization schemes: Pixel Norm (PN), Equalized Learning Rate (EQLR) and Spectral Normalization. For the first time, we show systematic comparison of the effects of each on a generative model. By improving the loss function with a margin term, we are able to drop PN and EQLR while solving the stability problem.

{\it (iii)}~With these improvements, our model can reconstruct $256{\times}256$ \CelebAHQ images with substantially better conservation of identity and attribute editability than shown for the baseline models such as the state-of-the-art model IntroVAE, with almost comparable sample quality. In comparison to the vanilla \PIONEER, we show substantially better conservation of face identity (via L2 distance in the embedding space of a face recognition model \cite{dlib09,ageitgey}) and significantly improved LPIPS \cite{zhang2018} and FID \cite{heusel2017} scores. In \LSUN Bedrooms dataset, we improve the disentanglement performance of the vanilla \PIONEER despite our model being simpler.

\section{Related work}
\label{sec:related}
\noindent
Our work builds upon the previous research in generative image models, such as variational autoencoders (VAEs, \cite{kingma2014,rezende2014}), autoregressive models, flow models, and GAN variants. Our approach borrows ideas from both autoencoders and GANs. In this section, we give a brief overview of the background literature and conclude with discussion about the closest related works.

The basic idea of a GAN is to train two networks, generator and discriminator, in a competitive manner so that \emph{(a)}~the generator learns to produce images from the same distribution as the training data  and \emph{(b)}~the discriminator learns to distinguish the synthetic images produced by the generator from the real training samples as well as possible. If successful, the generator has become good enough so that the discriminator can no longer make the distinction. The generator starts from a compact random latent code.

During the recent years there has been rapid progress related to GAN-based models and applications. Many recent improvements improve the stability and robustness of the training process by suggesting new loss functions \cite{Arjovsky2017b}, regularization methods \cite{miyato2018,RothLNH17,mescheder2018}, multi-resolution training \cite{karras2017, zhang2016}, architectures \cite{karras2018style}, or combinations of these.

However, despite the notable progress in image generation, it is widely accepted that the capability for realistic image synthesis alone is not sufficient for most applications, such as image manipulation, where we start from an existing image. This calls for a model with an encoder, such as a VAE. On the other hand, autoencoders tend to suffer from blurry outputs, and are often used for only low resolution images (even in recent works such as \eg \cite{tabor2018,tolstikhin2018,qichen2018}, see also comparison in \cite{heljakka2018}).

Thus, there have been many efforts to combine GANs with autoencoder models,  (\eg \cite{brock2016,Rosca2017,Donahue2017,dumoulin2016D}). For instance, \cite{Donahue2017} and \cite{dumoulin2016D} proposed utilizing three deep networks in order to learn functions that enable mappings between the data space and the latent space in both directions. That is, besides the typical autoencoder architecture, consisting of a decoder (\ie~generator) and encoder networks, their approach uses an additional discriminator network, which is trained to classify tuples of image samples with their latent codes. Other authors introduce additional discriminator networks besides the generator and encoder. For example, \cite{larsen2015,brock2016} use a GAN-like discriminator in sample space and \cite{makhzani2015,mescheder2017} in latent space. Nevertheless, the image synthesis performance of these hybrid models has not yet been shown to match state-of-the-art of purely generative models \cite{karras2017, karras2018style}.

In this paper, we build upon \PIONEER \cite{heljakka2018}, based on the adversarial generator--encoder (AGE) \cite{ulyanov2017}. In contrast to many other previous works, these two models consist of only two deep networks, a generator and an encoder, which represent the mappings between the image space and latent space. In addition, the method of progressive network growing, adapted from \cite{karras2017}, is utilized in \cite{heljakka2018}.

The results of \cite{heljakka2018} are promising, and both synthesis and reconstruction have good quality in relatively high image resolutions. However, in this paper, we show that \cite{heljakka2018} suffers from large fluctuations of the competing divergence terms of the adversarial loss, and this seems to hamper optimization and convergence thereby limiting performance.

Besides \cite{heljakka2018}, other recent and related works are IntroVAE \cite{huang2018} and GLOW \cite{Kingma+Dhariwal:2018}. Based on VAE, IntroVAE is fundamentally different from \PIONEER that is based on AGE. IntroVAE has been shown to produce high quality samples, but not to faithfully conserve sample details such as identity of faces (see Fig.~3 in \cite{huang2018}). Mere conservation of overall image topology is generally insufficient for manipulating image attributes. Based on the contributions of our paper, we are able to show that AGE-based generative models are capable of producing competitive results at $256{\times}256$ resolution, while conserving the face identity better than IntroVAE. This is in contrast to the observations in \cite{huang2018}, where the authors were not able to make AGE training converge with large image resolutions. This finding is particularly promising since the \PIONEER model has a simpler yet more powerful architecture than the corresponding purely generative model PGGAN \cite{karras2017}. It shows that the conventional GAN paradigm of a separate discriminator network is not necessary for learning to infer and generate image data sets.

Our model learns to manipulate image attributes in a fully unsupervised manner, in contrast to supervised approaches where the class information is provided during training (\eg \cite{lample2017, heljakka2018a}), and to models only capable of specific discrete domain transformations \cite{zhu2017,kim2017,yi2017,choi2017,isola2016}. In practice, all prior unsupervided work uses $64{\times}64$ resolution, such as every model cited in a recent large-scale autoencoder comparison \cite{locatello2018}. In the GAN research line, the state-of-the-art models such as \cite{karras2018style} could be used for this in high resolution, but they have no encoder to deal with new input images.

\section{Methods}
\label{sec:methods}
\noindent
In this section, we start from the basics of \PIONEER \cite{heljakka2018} which in turn builds on \cite{ulyanov2017} and \cite{karras2017}  (Sec.~\ref{sec:training}). From the perspective of image manipulation in $256{\times}256$ resolution (and above), the regular \PIONEER reaches reasonable sample quality and diversity, but the reconstructions are often not faithful to the originals (unlike in $128{\times}128$ \CelebA).

Therefore, our main task in this paper is to substantially improve the reconstructions without sacrificing other performance aspects. For this, we must address two issues:
{\em(i)}~On the training behavior level, the \PIONEER training dynamics are difficult to optimize, due to wildly oscillating Kullback--Leibler divergence (KL) terms during training. Further, occasionally the encoder and decoder completely diverge and the training performance collapses (Fig.~\ref{fig:competing-a}).
{\em(ii)}~On the algorithmic level, the three different normalization schemes used in baseline \PIONEER compound the difficulty of understanding the behavior of the model, and together appear to contribute to the issue (1) in complicated ways.
Instead of proposing yet another network architecture, we seek to find the minimum sufficient changes to simplify and stabilise the training workflow.

We benefit from first simplifying the algorithm, and reduce the three normalization schemes to only one (Sec.~\ref{sec:simplification}).
This simplification comes at the cost of even less stable training dynamics.
However, the behavior becomes more straightforward, and we are able to stabilize the training by bounding the difference between the competing KL terms, preventing them from diverging (Sec.~\ref{sec:competing}). We also compare the impact of each combination of normalization strategies on early training stages (Fig.~\ref{fig:competing-a}).

Finally, we analyze the representational power of our improved model by measuring the degree of disentanglement in the model's latent space and demonstrate realistic image feature manipulations (Sec.~\ref{sec:disentanglement}).

\subsection{Model training dynamics}
\label{sec:training}
\noindent
In \PIONEER, the encoder $\phi$ and decoder $\theta$ are trained separately during each training step, with separate loss functions. Because of this separation and the presence of loss terms that are the exact opposites of each other, the training can be called adversarial. With $\vx \sim X$ as the training samples and $\hat{\vx}$ = $\vtheta(\hat{\vz})$ where $\hat{\vz} \sim \mathrm{N}(\vect{0},\MI)$, the encoder loss consists of the encoder trying to push the distribution of the latent codes of the training samples $q_\phi(\vz\mid\vx)$ towards a unit Gaussian distribution $\mathrm{N}(\vect{0},\MI)$ and the distribution of the codes of the generated samples $q_\phi(\vz\mid\hat{\vx})$ away from $\mathrm{N}(\vect{0},\MI)$. The decoder tries to do the opposite to the generated samples. Furthermore, the encoder attempts to minimize reconstruction error $L_\mathcal{X}$ with L1 distance in sample space $\mathcal{X}$, and the decoder to minimize reconstruction error $L_\mathcal{Z}$ with cosine distance in latent code space $\mathcal{Z}$. Hence, as explained in detail in \cite{heljakka2018}, the full loss functions are as follows:
\begin{align}
  L_\phi =~&\mathrm{D_{KL}}[q_\phi(\vz\mid\vx) \,\|\, \mathrm{N}(\vect{0},\MI)] \nonumber \\ -~&\mathrm{D_{KL}}[q_\phi(\vz\mid\hat{\vx}) \,\|\, \mathrm{N}(\vect{0},\MI)] + \lambda_\mathcal{X} L_\mathcal{X}, \label{eq:loss-phi} \\
  L_\theta =~&\mathrm{D_{KL}}[q_\phi(\vz\mid\hat{\vx}) \,\|\, \mathrm{N}(\vect{0},\MI)] + \lambda_\mathcal{Z}L_\mathcal{Z}, \label{eq:loss-theta}
\end{align}
with $D_\mathrm{KL}$ denoting Kullback--Leibler divergence. The reconstruction loss terms are defined as:
\begin{align}
  L_\mathcal{X}(\vtheta,\vphi) &= \mathbb{E}_{\vect{x} \sim X} \| \vect{x} -\vtheta(\vphi(\vect{x})) \|_1, \\
  L_\mathcal{Z}(\vtheta,\vphi) &= \mathbb{E}_{\vect{z} \sim \mathrm{N}(\vect{0},\MI)}{[1 - \vect{z}\T \vphi(\vtheta(\vect{z}))]}.
\end{align}
Vectors in $\mathcal{Z}$ are always normalized to unity.
Contrasted to regular autoencoders, the second major aspect of \PIONEER is the progressive growing of the network architecture during training, as follows. The encoder and decoder are divided into residual convolution/deconvolution blocks operating on separate resolutions ($16{\times}16$, $32{\times}32$, \ldots). The training is divided into phases during which we only train the blocks that operate up to that resolution. Once that resolution is sufficiently learnt, we gradually fade in the next level blocks, and so on. (For more details, see \cite{heljakka2018,karras2017}.)

When training on resolutions $128{\times}128$ or higher, we observe that around the time when the image generation results begin to get worse, the encoder tends to assign increasingly low KL divergence estimates for the training samples, and increasingly large ones for the generated samples. In other words, the encoder wins the game. We will keep this observation in mind, and now turn to the weight normalization schemes that underlie the training dynamics.

\subsection{Simplification of the normalization scheme}
\label{sec:simplification}
\noindent
In \cite{heljakka2018}, three unrelated implicit regularization techniques are used to stabilize the training: {\em(i)} equalized learning rate \cite{karras2017} and {\em(ii)} pixel norm \cite{karras2017} in the decoder, as well as {\em(iii)}~spectral normalization \cite{miyato2018} in the encoder. (i) is used to scale generator weights by dividing each weight with He's initializer \cite{he2015} at runtime. (ii) normalizes, for each convolutional layer, the feature vector of each pixel to unit length. In (iii), the spectral norm of each layer of the network is constrained directly during every computation pass by dividing each network weight matrix by its largest singular value. This keeps the Lipschitz constant under control. Intuitively, (iii) ensures that the weights do not amplify the `scale' of the input signal, but allows them to freely `rotate' the signal.

Since (i) scales the weights based on the number of input connections, it merely ensures that the overall scales of weight gradient updates are balanced against the changing layer sizes. While this helps to even out extreme cases, it does not guarantee Lipschitz continuity. 

Method (ii) scales the outputs of the convolution operations. First, it prevents activations from escalating to large values. Second, it eliminates much of the information about how the activation of a given filter varies across locations. The filter might activate more strongly on one pixel than on another, but this difference may be `scaled away' depending on the other filter activations.
This explains the observed dampening effect that (i) and (ii) have on learning, with more stability at the cost of capacity.

\begin{figure*}[!t]
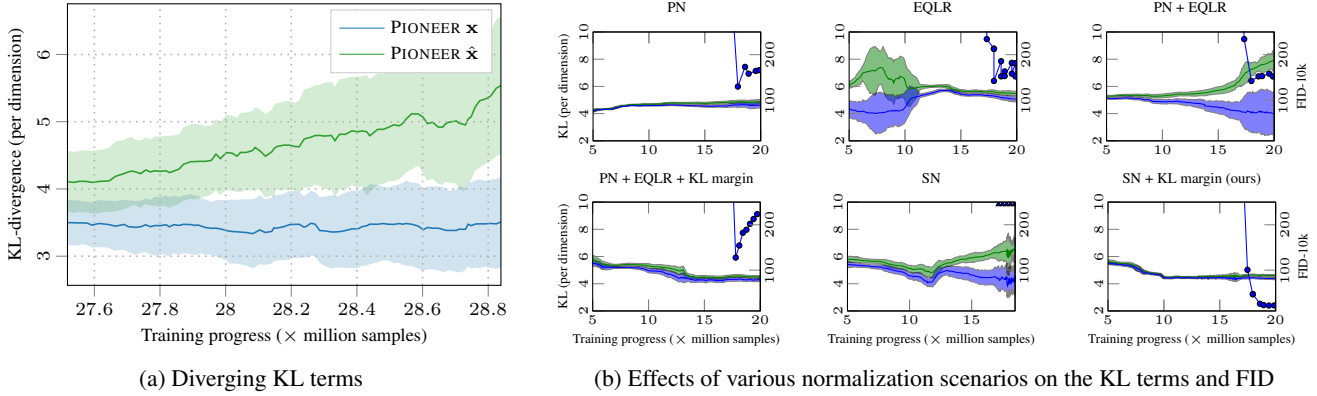

  \centering\scriptsize
  \setlength{\figurewidth}{.33\textwidth}
  \setlength{\figureheight}{.65\figurewidth}
  \pgfplotsset{yticklabel style={rotate=90}, ylabel near ticks, clip=true,scale only axis,axis on top,clip marker paths,legend style={row sep=0pt},xlabel near ticks,legend style={fill=white}}
  \begin{subfigure}[b]{.38\textwidth}
 	\input{./fig/fig4/fig4a-arno.tex}
 	\vspace*{-1.2em}
 	\caption{Diverging KL terms}
 	\label{fig:competing-a}
  \end{subfigure}
  \hfill
  \begin{subfigure}[b]{.58\textwidth}
    \tiny
    \setlength{\figurewidth}{.22\textwidth}
    \setlength{\figureheight}{.65\figurewidth}
    \pgfplotsset{set layers,scale only axis}
    \begin{minipage}[c][.22\textwidth][t]{.3\textwidth}
      \centering
      \input{./fig/fig3b/fig3b1.tex}
    \end{minipage}
    \hfill
    \begin{minipage}[c][.22\textwidth][t]{.3\textwidth}
      \centering
      \input{./fig/fig3b/fig3b2.tex}
    \end{minipage}
    \hfill
    \begin{minipage}[c][.22\textwidth][t]{.3\textwidth}
      \centering
      \input{./fig/fig3b/fig3b3.tex}
    \end{minipage}\\[0em]
    \begin{minipage}[c][.22\textwidth][t]{.3\textwidth}
      \centering
      \input{./fig/fig3b/fig3b4.tex}
    \end{minipage}
    \hfill
    \begin{minipage}[c][.22\textwidth][t]{.3\textwidth}
      \centering
      \input{./fig/fig3b/fig3b5.tex}
    \end{minipage}
    \hfill
    \begin{minipage}[c][.22\textwidth][t]{.3\textwidth}
      \centering
      \input{./fig/fig3b/fig3b6.tex}
    \end{minipage}
    \vspace*{1em}
 	\caption{Effects of various normalization scenarios on the KL terms and FID}
 	\label{fig:competing-b}
  \end{subfigure} 	
  \caption{(a) The competing KL divergence terms that the encoder assigns for $\vx$ (training samples) and $\hat{\vx}$ (generated samples) in \PIONEER show two pathological properties. Relatively early on, the encoder overpowers the decoder leading the terms to diverge from each other, and the dynamic range of each grows, leading to an upper bound on the learning capacity. (b) The individual contribution of our changes, shown for KL divergence terms for early stages of \CelebAHQ training. FID (blue dots) is shown for $128 {\times} 128$ stage onwards (with stage fade-in starting at 16.52M samples). PixelNorm (PN) or Equalized Learning Rate (EQLR) alone or in combination (top) will lead to poor FID even with KL margin (bottom left). On the other hand, merely adding Spetral Normalization (SN) without KL margin will lead to divergence of the terms and training collapse (bottom middle). Adding SN with the margin produces the optimal outcome (bottom right).}

  \label{fig:competing}
  \vspace*{-.5em}
\end{figure*}

Informally, we can construe the regularization as having two goals: {\em (A)}~stable individual training steps and {\em (B)}~preventing the encoder--decoder competition from spiralling out of control. (i) and (ii) meet (A) but fail (B). As their effect is local, they provide nothing to prevent the encoder from over-powering the decoder over time (Fig.~\ref{fig:competing}). Likewise, (iii) guides only the local learning, but with less dampening.

As (iii) is more principled an approach than (i) and (ii), we could in fact use it to replace (i) and (ii) altogether. Hypothetically, without the strong dampening effect, this buys us more learning capacity. Empirically, however, it then leads to unbalanced competition even faster than with (i) and (ii).

Now, the key insight here is to use the implicit regularizer (iii) to address goal (A) {\it only}. To this end, the weaker constraints of (iii) will suffice for encoder and decoder alike, allowing us to discard (i) and (ii) altogether. For (B), we will proceed to explicitly improve the loss function, instead.

\subsection{Competing KL divergence terms}
\label{sec:competing}
\noindent
We now return to the KL terms. In general, their absolute values do not seem to always correlate with better generation results. Instead, the difference between them is critical.

We thus set out to regularize this part of the training by adding a simple hinge loss to limit the gradient reward for the encoder. A major component of the gradient of the encoder comes from the gap between the KL terms. We wish to ensure that the encoder is not motivated to increase the gap too much, so we define a single margin term $M_\mathrm{gap}$ that defines the upper bound of the gap, and then modify Eq.~\eqref{eq:loss-phi} as follows into a hinge-loss form:
\begin{multline}
  L_\phi = \max(-M_\mathrm{gap}, \mathrm{D_{KL}}[q_\phi(\vz\mid\vx) \,\|\, \mathrm{N}(\vect{0},\MI)] \\ \mathrm{D_{KL}}[q_\phi(\vz\mid\hat{\vx}) \,\|\, \mathrm{N}(\vect{0},\MI)]) + \lambda_\mathcal{X} L_\mathcal{X}. \label{eq:loss-phi-mod}
\end{multline}
Because the latter KL term is, in practice, almost always larger than the former, the gap is negative and needs to be bounded from below. Now, Eq.~\eqref{eq:loss-phi-mod} alone would not care about the absolute values of the KL terms at all. However, since Eq.~\eqref{eq:loss-theta} remains unchanged, the decoder training provides the force that drives the $\mathrm{D_{KL}}[q_\phi(\vz\mid\hat{\vx}) \,\|\, \mathrm{N}(\vect{0},\MI)]$ lower. Combined with Eq.~\eqref{eq:loss-phi-mod}, the result is a force that pushes the other KL term lower, too (see Fig.~\ref{fig:competing} for an example of diverging KLs and the effect of bounding the gap).

After applying this change, the training becomes stable again, regardless of resolution, while using our simplified weight normalization scheme. There is no collapse of the training, and furthermore, we tend to see steady improvements well beyond the capacity of the original \PIONEER. With the approach in Eq.~\eqref{eq:loss-phi-mod}, we do not need to try to reduce the learning rate to dampen the gradients.
We note that the large gradients are not a problem in the early pre-training stages. On the contrary, applying a heavy-handed margin in the early stages will reduce the rate of learning too much. Hence, it is sufficient to only apply the margin after the pre-training. Alternatively, we could have defined the margin to be dependent on the progressive stage of the training, but more experiments should be done to confirm the generality of such margin choices.

\subsection{Disentanglement of latent space}
\label{sec:disentanglement}
\noindent
One of many ways to measure disentanglement of the latent space is to find latent directions that correspond to specific factors of variation in the sample space.
In an unsupervised training setup, these factors are unknown {\it a~priori}. However, as a proxy measure, we can measure how smoothly the generated samples change while we move around in the latent space. This requires a good metric (inductive bias) for measuring such changes between images, such as LPIPS \cite{zhang2018}. Following \cite{karras2018style}, we compute the Perceptual Path Length (PPL) by repeatedly taking a short constant-length random vector in the latent space, generating images at its endpoints, and measuring LPIPS between the points.

Visually, we can modify isolated image attributes, such as the degree of smiling, by finding the corresponding latent code vectors after the training. We take our existing model \textit{trained without labels}, a set $A$ of images with the desired attribute, and a set $B$ without it. \textit{Requiring no additional training or optimization}, we take the difference between the mean latent code of each set. The resulting 512-dimensional difference vector can then be added to the code of any new real test image $\vect{x}_\mathrm{noF}$, scaled to the desired intensity $\lambda$. The decoded image then gains (or, with negative $\lambda$, loses) the attribute (see Fig.~\ref{fig:manipulate}):
\begin{equation*}
  \vect{x}_\mathrm{F}(A,B) = 
  \vtheta(\mathrm{\vphi(\vect{x}_\mathrm{noF})} + \lambda [\mathbb{E}_{\vect{x} \sim A} \| \vphi(\vect{x}) \|  - \mathbb{E}_{\vect{x} \sim B} \| \vphi(\vect{x}) \|]).
\end{equation*}

\begin{figure}
  \centering\scriptsize
  \setlength{\figurewidth}{.4\textwidth}
  \setlength{\figureheight}{.6\figurewidth}
  \pgfplotsset{yticklabel style={rotate=90}, ylabel near ticks,clip=true,scale only axis,axis on top,clip marker paths,legend style={row sep=0pt},xlabel near ticks,legend style={fill=white}}
\begin{tikzpicture}

\definecolor{color0}{rgb}{0.12156862745098,0.466666666666667,0.705882352941177}
\definecolor{color1}{rgb}{0.172549019607843,0.627450980392157,0.172549019607843}

\begin{axis}[
height=\figureheight,
legend cell align={left},
legend entries={{\PIONEER},{Balanced \PIONEER}},
legend style={draw=white!80.0!black},
tick align=outside,
tick pos=left,
width=\figurewidth,
x grid style={white!69.01960784313725!black},
xlabel={Training progress ($\times$ million samples)},
xmin=21.481, xmax=29.819,
y grid style={white!69.01960784313725!black},
ylabel={Frech{\'e}t Inception Distance (FID-10k)},
ymin=24.8773118214624, ymax=74.5864462988982
]
\addplot [semithick, color0, mark=*, mark size=3, mark options={solid}, only marks, draw=black]
table [row sep=\\]{%
23.16	67.886262517824 \\
23.8	70.6133869855567 \\
24.6	62.3383290069143 \\
25	49.5875850177829 \\
25.56	72.3269401862875 \\
25.72	56.3706932686147 \\
25.8	63.4467131273834 \\
26.22	46.0892319517875 \\
26.78	44.3769211184193 \\
27.2	44.4407726431373 \\
27.48	40.9328802764307 \\
27.76	42.5604872574268 \\
28.04	42.5417438519239 \\
28.18	48.9041464141129 \\
28.32	52.3002891960483 \\
28.88	45.2443910064547 \\
29.44	48.8701848391476 \\
};
\addplot [semithick, color1, mark=square*, mark size=3, mark options={solid}, only marks, draw=black]
table [row sep=\\]{%
21.86	33.4381615188779 \\
22.84	27.3546102961532 \\
23.08	37.4946950942013 \\
23.4	28.6148111589679 \\
24.1	29.1586219109654 \\
24.52	28.4001532119065 \\
24.68	29.5870762336461 \\
25.48	27.2613561343609 \\
26.92	27.1368179340732 \\
27.4	29.8711912630575 \\
27.56	29.7013123732603 \\
27.88	29.0320679322418 \\
};
\end{axis}

\end{tikzpicture}
  \caption{FID-10k values of \CelebAHQ during training, with regular \PIONEER and Balanced \PIONEER. The FID measures the quality and diversity of generated images (smaller is better). The regular \PIONEER scores are markedly worse and more volatile.}
  \label{fig:fid}
\end{figure}
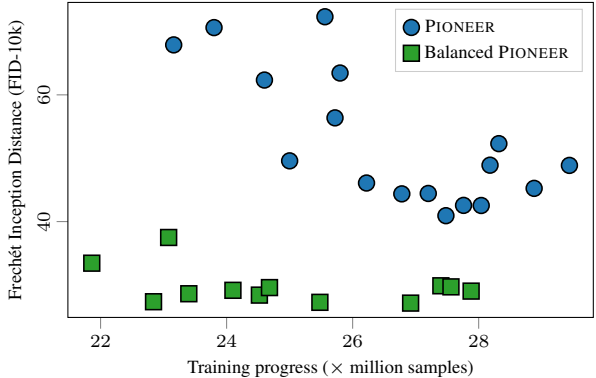

\section{Experiments}
\label{sec:experiments}
\noindent
Any generative models can create data samples, but encoder--decoder models can also take input samples and modify them via latent space. Correspondingly, one can test such a model by reconstructing new test samples and interpolating between them in the latent space, evaluating some generated random samples, and analyzing the latent space.

The various ways of measuring generative models constitute an active field of research as such. To evaluate random samples, we look for both quality and diversity of the sampled distribution, typically via Fr\'echet inception distance (FID, \cite{heusel2017}) or Inception Distance \cite{Salimans2016}. They come with various shortcomings \cite{binkowski2018} but as long as we use identical sample size, FID remains a reasonably reliable tool for comparing different models. For easy comparison to prior work, we measure against the training set (but see \cite{kurach2018} for limitations).

\begin{table}[!tb]
  \caption{Comparison of Fr\'echet Inception Distance (FID) and perceptual path length (PPL) of $256{\times}256$ images on \CelebAHQ and \LSUN dataset between Balanced \PIONEER, regular \PIONEER, PGGAN, IntroVAE and GLOW.  Balanced \PIONEER has the best PPL result, PGGAN has the best FID results. Improvement of the Balanced \PIONEER over the regular \PIONEER is clear on \CelebAHQ. \cite{huang2018} provides the $256{\times}256$ FID of IntroVAE for \LSUN Bedrooms but not for \CelebAHQ (they provide it for $1024{\times}1024$ as 5.19, but with a more favorable train/test split). For other figures, the best-run results are shown. For GLOW, $256 {\times} 256$ \LSUN image generation has not been demonstrated (only $128 {\times} 128$). FID is based on a 50k batch of generated samples compared to training samples. \CelebAHQ Perceptual Path Length (PPL) was calculated with 100k samples, cropped to $128{\times}128$ ($\varepsilon = 10^{-4}$), \LSUN PPL with $256{\times}256$. Pre-trained models for PGGAN and GLOW ($T=0.9$ resulted in best FID) were used with default settings provided by the authors. For all numbers, \textbf{smaller is better}.}
  \vspace*{3pt}
  \label{tbl:results}
  \centering
  {\footnotesize\noindent\scriptsize%
  \setlength{\tabcolsep}{0pt}
  \begin{tabular*}{\columnwidth}{@{\extracolsep{\fill}} lcccccc}
  \toprule
    & FID  & FID & PPL & PPL \\
    & (\CelebAHQ) & (\LSUN) & (\CelebAHQ) & (\LSUN)\\
  \midrule
  PGGAN  & $\bf 8.03 \bf$ & $\bf8.34 \bf$ & $229.2$ & $1080.1$\\
  IntroVAE  & --- & $8.84$ & --- & ---\\
  GLOW  & $68.93$ & --- & $219.6$ & ---\\
  \PIONEER & {{$39.17$}} & {$ 18.07 $}  & 155.2 & 779.5\\  
  Balanced \PIONEER (ours) &   {$25.25$} & $17.89$  & $\bf 146.2 \bf$ & $\bf 678.4 \bf$\\
  \bottomrule
  \end{tabular*}}
\end{table}

\begin{figure*}[!t]
  \centering\scriptsize
  \setlength{\figurewidth}{0.081\textwidth}
  \setlength{\figureheight}{\figurewidth}

  \begin{subfigure}{\textwidth}
  \begin{tikzpicture}[inner sep=0]

    \foreach \x [count=\i] in {57,55,56,54,58,59,60,61,63}
      \node[draw=white,fill=black!20,minimum size=\figurewidth,inner sep=0pt]
        (\i) at ({\figurewidth*mod(\i-1,3)},{-\figureheight*int((\i-1)/3)})
        {\includegraphics[width=\figurewidth]{./fig/fig_random/pine/25480001_9\x.jpg}};

    \foreach \x [count=\i] in {0,...,8}
      \node[draw=white,fill=black!20,minimum size=\figurewidth,inner sep=0pt]
        (\i) at ({3.1*\figurewidth+\figurewidth*mod(\i-1,3)},{-\figureheight*int((\i-1)/3)})
        {\includegraphics[width=\figurewidth]{./fig/fig_random/glow_07/rand071\x.jpg}};                

    \foreach \x [count=\i] in {0,...,8}
      \node[draw=white,fill=black!20,minimum size=\figurewidth,inner sep=0pt]
        (\i) at ({6.2*\figurewidth+\figurewidth*mod(\i-1,3)},{-\figureheight*int((\i-1)/3)})
        {\includegraphics[width=\figurewidth]{./fig/fig_random/pggan/015-pgan-celeba-preset-v2-4gpus-fp32-network-snapshot-010000-00000\x.jpg}};

    \foreach \x [count=\i] in {1,...,9}
      \node[draw=white,fill=black!20,minimum size=\figurewidth,inner sep=0pt]
        (\i) at ({9.3*\figurewidth+\figurewidth*mod(\i-1,3)},{-\figureheight*int((\i-1)/3)})
        {\includegraphics[width=\figurewidth]{./fig/fig_random/introvae_random/intro\x.jpg}};

    \node at ({\figurewidth},{-2.7*\figureheight}) {Balanced \PIONEER (ours)};
    \node at ({3.1*\figurewidth+\figurewidth},{-2.7*\figureheight}) {GLOW};
    \node at ({6.2*\figurewidth+\figurewidth},{-2.7*\figureheight}) {PGGAN};
    \node at ({9.3*\figurewidth+\figurewidth},{-2.7*\figureheight}) {IntroVAE};
        
  \end{tikzpicture}
  \caption{\CelebAHQ $256{\times}256$}
  \label{fig:celeba}
  \end{subfigure}\\[1em]

  \begin{subfigure}{\textwidth}
  \begin{tikzpicture}[inner sep=0]

    \foreach \x [count=\i] in {30,...,38}
      \node[draw=white,fill=black!20,minimum size=\figurewidth,inner sep=0pt]
        (\i) at ({\figurewidth*mod(\i-1,3)},{-\figureheight*int((\i-1)/3)})
        {\includegraphics[width=\figurewidth]{./fig/fig_random/lsun/pine/lsun25400/25400001_73\x.jpg}};

    \foreach \x [count=\i] in {56,...,64}
      \node[draw=white,fill=black!20,minimum size=\figurewidth,inner sep=0pt]
        (\i) at ({3.1*\figurewidth+\figurewidth*mod(\i-1,3)},{-\figureheight*int((\i-1)/3)})
        {\includegraphics[width=\figurewidth]{./fig/fig_random/lsun/pine_Old/B27384001_9\x.jpg}};                

    \foreach \x [count=\i] in {0,...,8}
      \node[draw=white,fill=black!20,minimum size=\figurewidth,inner sep=0pt]
        (\i) at ({6.2*\figurewidth+\figurewidth*mod(\i-1,3)},{-\figureheight*int((\i-1)/3)})
        {\includegraphics[width=\figurewidth]{./fig/fig_random/lsun/pggan/00000\x.jpg}};

    \node at ({9.3*\figurewidth+\figurewidth},{-\figureheight}) {\includegraphics[width=3\figurewidth]{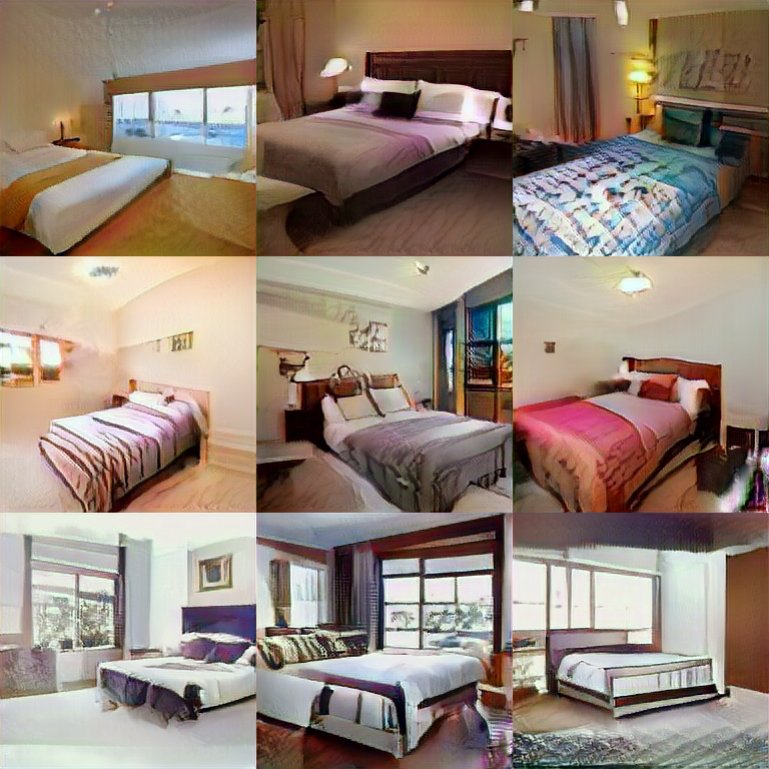}};

    \node at ({\figurewidth},{-2.7*\figureheight}) {Balanced \PIONEER (ours)};
    \node at ({3.1*\figurewidth+\figurewidth},{-2.7*\figureheight}) {\PIONEER};
    \node at ({6.2*\figurewidth+\figurewidth},{-2.7*\figureheight}) {PGGAN};
    \node at ({9.3*\figurewidth+\figurewidth},{-2.7*\figureheight}) {IntroVAE};
        
  \end{tikzpicture}
  \caption{LSUN Bedrooms $256{\times}256$}
  \label{fig:lsun}
  \end{subfigure}\\[1em]
  \caption{Generated samples from \CelebAHQ and LSUN Bedrooms. IntroVAE from \cite{huang2018} (\CelebAHQ downscaled), others uncurated.}
  \label{fig:celaba-lsun}
  \vspace*{-.5em}
\end{figure*}

To measure the extent to which face identity of an input image is conserved in the reconstructed face images, we used L2 distance between the 128-D feature embeddings of the two images, in the embedding space of a pre-trained DLib face recognition model \cite{dlib09,ageitgey}. To measure the conservation of overall image features, we use LPIPS \cite{zhang2018} which correlates with human judgement more closely than, for example, the commonly used pixel space L2 distance or Structural Similarity measures \cite{Zhou+Bovik+Sheikh+Simoncelli:2004}. Finally, to evaluate the quality of the latent space as such, we measure the Perceptual Path Length \cite{karras2018style} and show interpolations between reconstructions of input images as well as feature modifications. The success in the latter requires a well-structured latent space.

Since \PIONEER networks are most useful in the domain of high-resolution images ($128{\times }128$ and higher), we consider the \CelebAHQ \cite{karras2017} and \LSUN Bedrooms \cite{Yu:2015} datasets with images up to $1024{\times}1024$ and $256{\times}256$ resolution, respectively. \CelebAHQ contains 30,000 images (where we use 27,000 / 3,000 split for training and testing images) while \LSUN has its designated separate testing images. In Table~\ref{tbl:results}, we compare the \PIONEER, Balanced \PIONEER, PGGAN and GLOW, with both datasets, in $256{\times}256$.

We train progressively as in \cite{heljakka2018}, allowing the last two pre-training phases ($64{\times}64$ and $128{\times}128$) last 0.5--2 times longer than the earlier phases. The final phase is trained until convergence of the FID metric.
We train both the \CelebAHQ model and the \LSUN Bedrooms model for 8~days on two Nvidia V100 GPUs, up to 25.5M samples (resolutions vary between epochs). In higher resolutions, the batch size is reduced. After the pre-training stages up to $64{\times}64$, we switch on the margin ($m=0.6$ for \LSUN, $m=0.2$ for \CelebAHQ). From $256{\times}256$ onwards, $m=0.4$. The fine-tuning is not necessary, but slightly improves the FID values. The overall scale of $m$ can be determined empirically from the behavior of KL divergence values.

\subsection{Ablation studies in varying the normalization}
\noindent
We carried out a sequence of ablation studies to quantify each single change in the normalization scheme. Fig.~\ref{fig:competing-b} focuses on the critical early training stages separately using PixelNorm (PN), Equalized Learning Rate (EQLR), PN+EQLR \etc\ in \CelebAHQ up to 20M steps. Other hyper-parameters and resolution schedule remain the same. PN or EQLR alone or in combination will lead to poor FID even with KL margin. On the other hand, Spetral Normalization (SN) without KL margin will lead to divergence of the KL terms and training collapse. Only adding SN with the margin produces the optimal outcome. Also in PGGAN, replacing PN+EQLR with SN retained the FID results with \CelebAHQ in $128{\times}128$ (see Supplement).

\begin{figure*}[!t]
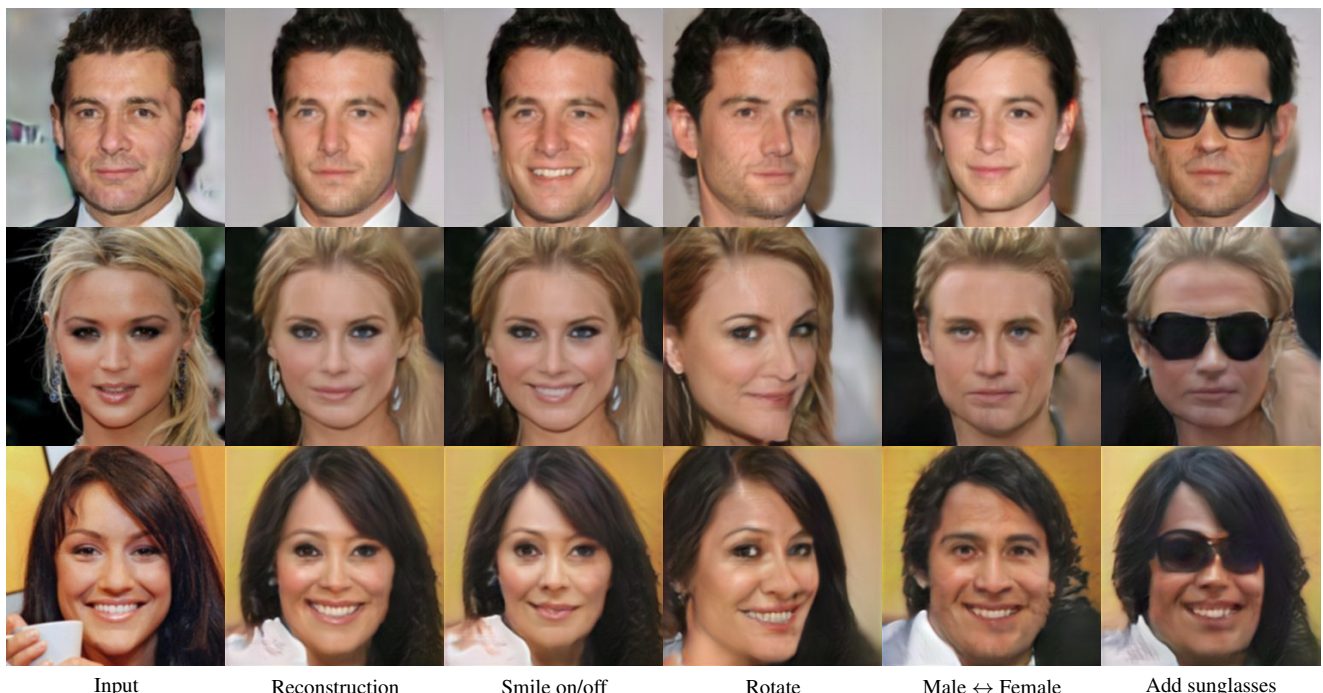

  \centering\footnotesize
  \setlength{\figurewidth}{.1667\textwidth}
  \setlength{\figureheight}{\figurewidth}

  \begin{tikzpicture}[inner sep=0]

  \tikzstyle{arrow} = [draw=black!10, single arrow, minimum height=10mm, minimum width=3mm, single arrow head extend=1mm, fill=black!10, anchor=center, rotate=0, inner sep=2pt]

  \foreach \x [count=\i] in {0,1,2,4,3,5} {
     \node[] at ({\figurewidth*\i},{0*\figureheight}) {\includegraphics[width=\figurewidth]{./fig/fig_manip/10\x.jpg}};
     \node[] at ({\figurewidth*\i},{-1\figureheight}) {\includegraphics[width=\figurewidth]{./fig/fig_manip/20\x.jpg}};
     \node[] at ({\figurewidth*\i},{-2\figureheight}) {\includegraphics[width=\figurewidth]{./fig/fig_manip/30\x.jpg}};
     }

  \foreach \x [count=\i] in {Input,Reconstruction,Smile on/off,Rotate,Male $\leftrightarrow$ Female,Add sunglasses} {  
     \node[fill=white,inner sep=0, minimum width=4mm] at ({\figurewidth*\i},-2.6\figureheight) {\x};
     
     }

  \end{tikzpicture}
  \caption{Examples of feature manipulation in $256{\times}256$ resolution \CelebAHQ test set images that the model has not seen before, for attributes that the model did not know about during training. Column~1: the input; Column~2: the reconstruction; Columns~3--6: various added feature vectors. Each 512-dimensional feature vector was extracted after training as a simple arithmetical difference between the latent vector of 32 to 64 training set examples that contain and do not contain the feature. The difference vector was then added to the latent code of each new input image and decoded back into a result image. There is no \textit{a~priori} reason to assume that this would result in a valid image in the first place, nor that the feature would be transformed. Note that there is no class information in the training set, which means that during training, the model has to figure out the existence of these features from scratch. Only a well-structured latent space can explain the result. Here, the feature intensity scale $\lambda$ varies between $[-2.0,2.0]$. For smoothly increasing the intensity, please see the Supplement.}
  \label{fig:manipulate}
\end{figure*}

\subsection{LSUN Bedrooms}
\noindent
For \LSUN Bedrooms, we show randomly generated samples in Fig.~\ref{fig:lsun} in $256{\times}256$. In terms of visual comparison to, for example, \cite{huang2018}, \cite{karras2017}, or \cite{gulrajani2017}, it is hard to discern the differences between the models. However, comparing to  $128{\times}128$ samples of \cite{Kingma+Dhariwal:2018} or \cite{heljakka2018} (which provide no $256{\times}256$ samples), we can already observe some mode collapse. We found that even the baseline \PIONEER can indeed generate high-quality samples of \LSUN Bedrooms at $256{\times}256$\footnote{This may be due to fixing the PyTorch bug 12671\cite{pyto12671} which prevented the spectral normalization from working on multi-GPU setup.}. The balanced \PIONEER reaches better PPL and similar FID figures with 30\% less training steps in the final stage than \PIONEER baseline (Table~\ref{tbl:results}). In terms of FID scores, our Balanced \PIONEER does not quite reach the state-of-the-art of \cite{karras2017} and \cite{huang2018}. However, the better performance of purely generative models such as \cite{karras2017} for random sampling is not directly comparable to that of encoder-based models.

\subsection{CelebA-HQ}
\noindent
In \cite{heljakka2018}, \PIONEER was shown to reconstruct \CelebA images up to $128{\times}128$, but only random samples at $256{\times}256$ level. We show reconstructions from both the baseline \PIONEER and Balanced \PIONEER at $256{\times}256$. Balanced \PIONEER has superior quality in both reconstructions (Fig.~\ref{fig:reconstructions}) and random sampling (Fig.~\ref{fig:celeba} and Table~\ref{tbl:results}). Reconstruction of input images is not interesting as such, but it serves as a proxy guarantee that the model at least handles the point in the latent space that represents the input image.

For identity conservation, the median L2 distance for 1k \CelebAHQ samples in the embedding space of a DLib face recognition model improved from 0.758 of the \PIONEER baseline to 0.712 of our model (29.1\% closer to the target of 0.600 at which the recognizer is confident that the person is same; see the Supplement for details). LPIPS score for 10k \CelebAHQ samples (cropped at face area to $128 {\times} 128$ for more precise measurement) improved from 0.223 of the baseline to 0.172 of our improved model (22.9\% reduction).

\subsection{Disentanglement measures}
\noindent
Finally, we evaluate whether the individual factors of variation in the training data are represented as disentangled directions in the latent space. We demonstrate this by showing smooth interpolations between the reconstructions of most input images (Fig.~\ref{fig:hex} and the Supplement) and attribute manipulation (Fig.~\ref{fig:manipulate}) in \CelebAHQ. We show superior degree of disentanglement with PPL, following \cite{karras2018style}. To this end, we sample 100k pairs of short random latent vector segments, the ends of which we decode into pairs of images, cropped around mid-face. The scaled expectation of their LPIPS value is, then, the PPL.

\section{Discussion and conclusions}
\label{sec:discussion}
\noindent
In this paper, we introduced a streamlined \PIONEER model variant, drawing from careful analysis of several normalization techniques and loss dynamics. Via state-of-the-art image manipulation capabilities, we also showed that AGE-based models can compete with GANs and VAEs.

We focused on face images to demonstrate the full range of capabilities of the model, even though it is in no way specifically designed for face data. Unlike GANs limited to random sample manipulation, our model allows for encoding, reconstruction and manipulation of new real inputs. Unlike VAEs, we showed sharp reconstruction and attribute editing up to $256{\times}256$ resolution. The reconstructive capacity and latent space disentanglement are superior to the results shown so far with vanilla \PIONEER and IntroVAE. Recent GANs and IntroVAE (but not GLOW) have shown better FID, but anything beyond random sample generation requires finer metrics.
To evaluate latent space structure, arguably the core of representation learning, we showed unsupervised editing of face attributes (also see the Supplement), sharp interpolations between new inputs, and PPL metrics superior to the baselines on both datasets (Table~\ref{tbl:results}).

The code for the experiments can be found at \url{https://github.com/AaltoVision/balanced-pioneer}.
\paragraph{Acknowledgements}
\label{sec:ack}
The authors acknowledge the Aalto Science-IT project, CSC -- IT Center for Science, Finland, and GenMind Ltd for computational resources, and support by the Academy of Finland grants 308640, 324345, 277685, and 295081.

\pagebreak

\phantomsection%
\addcontentsline{toc}{section}{References}

\begingroup
\bibliographystyle{ieee}
\interlinepenalty=10000
\bibliography{bibliography}
\endgroup

\clearpage
\appendix

\twocolumn[%
  \vspace*{2em}
  \begin{center}
  \Large\bf Supplementary material for\\
  Towards Photographic Image Manipulation with Balanced \\ Growing of Generative Autoencoders
  \end{center}
  \vspace*{5em}
]

\section{Training details}
\noindent
The architecture follows \cite{heljakka2018}, so that the encoder and the decoder are symmetric, composed of residual blocks with two convolutional layer each, and $1{\times}1$ convolution for a skip connection. The final layer of the encoder contains a 4x4 filter that reduces the feature map into 512 channels on a $1{\times}1$ map that, flattened, represents the latent vector. Other filters are $3{\times}3$. At each forward pass, the latent vector is normalized to unit length. Each block in the encoder halves the resolution of the input with stride equalling two, while each block of the decoder doubles it by upsampling and then running a convolution with stride equal to one.

The training proceeds via seven resolution phases, from $4{\times}4$ to $256{\times}256$. During the first half of each phase, the skip connection of the most recently added residual block is gradually faded out. During the second half, the skip connection is off. During each phase up to $32{\times}32$, the network sees a total of 2.4M image samples. The batch size generally halves for each consequtive phase to fit in GPU memory, with size 16 used for the final $256{\times}256$ stage. The pre-training phases $64{\times}64$ and $128{\times}128$ contained 3.5--6.2M samples so that the total sample count seen during pre-training for Balanced \PIONEER (20.04M in \CelebAHQ, 21.4M in \LSUN) was below the total for baseline \PIONEER. Two consequtive decoder training iterations were used for every single encoder training iteration.

Margin values in range $[0.2, 0.6]$ were tried. We switched on the margins after 13M training samples (\ie after the fade-in phase and the first full-length epoch of $64{\times}64$ phase). For the balanced \PIONEER, the results with the best FID reached for $256{\times}256$ resolution after seeing 27.3M training image samples were selected. This limit was chosen as it was the point around which the baseline \PIONEER reached best FID (10k) values for \LSUN Bedrooms, slightly earlier than its best value for \CelebAHQ (at 27.5M). Applying lower margin values, or even a margin of $0.2$ before $64{\times}64$ resolution stage, resulted in obvious training failure already by the end of the $64{\times}64$ stage. At the other extreme, applying very high margin values would nullify the effect of the margin.

Following \cite{heljakka2018,karras2017}, we maintain a moving exponential running average for the weights of the generator, and use it as the de facto generator after the training. Any other hyper-parameters follow \cite{heljakka2018}. Accordingly, every residual block in both the encoder and the decoder ends with a LeakyReLU (slope 0.2) activation function. However, at the final layer that maps into the latent vector, this activation in fact skews the distribution slightly away from the unit Gaussian, thus making the learning task harder. In follow-up works, we therefore strongly recommend removing the non-linearity of the last layer. We optimize with ADAM \cite{adam} ($\alpha=0.001, \beta_1 = 0$, $\beta_2 = 0.99$, and $\varepsilon=10^{-8}$).

\section{Ablation study details}
\label{sec:ablation}
\noindent
The comparisons in Fig.~\ref{fig:competing-b} were created by training the model with PN, PN+EQLR \etc\ in \CelebAHQ up to 20M steps. The schedule for resolution increases and other hyper-parameters except the normalization are the same as for the main experiments that produced the results reported for Balanced \PIONEER in the paper. For the experiments that involve the use of margin, we likewise follow the same schedule of margins as the main experiments. 
Finally, we tried removing all normalizations, and reproduced the negative results of \cite{heljakka2018}---the training simply fails to converge already before reaching $64 {\times} 64$ resolution. We also collect the FID results at 20M training steps for each method in the ablation study (Table \ref{tbl:results_abl}).

In addition, we modified the original PGGAN by replacing PN and EQLR in the generator with SN, and trained with \CelebAHQ in $128{\times}128$ up to 15,000,000 seen image samples. The best FID (measured with 10k training samples, 2 test runs) results were approximately the same (with PN+EQLR 12.44 and with SN 12.65). Note that here, there is no term corresponding to the KL margin.

\begin{table}[!tb]
  \caption{Comparison of Fr\'echet Inception Distance (FID) with various normalization schemes, after training with the first 20M samples in \CelebAHQ. The final FID (20M) and the best FID reached are shown. Note that some results could significantly improve with more training. 10,000 samples were used for FID, compared against the training set. Value for SN (without margin) is not given because the training consistenly collapses when reaching $128 {\times} 128$. For all numbers, \textbf{smaller is better}.}
  \vspace*{1em}
  \label{tbl:results_abl}
  \centering
  {\noindent\footnotesize
  \setlength{\tabcolsep}{0pt}
  \begin{tabular*}{0.75\columnwidth}{@{\extracolsep{\fill}} lcc}
  \toprule
  Method & FID-10k (20M) &  FID-10k (best)  \\
  \midrule
  PN  & $165.81 $ & $ 129.60 $\\
  PN+margin  & $ 223.90 $ & $ 117.00 $  \\
  EQLR  & $181.22$ & $162.99$  \\
  EQLR+margin  & $183.53$ & $ 128.97 $\\
  EQLR+PN  & $151.94$ & $ 140.60 $\\
  EQLR+PN+margin & $223.43$  & $127.65$ \\  
  SN &  ---  & ---   \\  
  SN+margin (ours) & {${ \bf 22.20} $} & {${ \bf 22.20} $}  \\  
  \bottomrule
  \end{tabular*}}
\end{table}

\section{Feature manipulation}
\begin{figure*}[h]
  \centering\scriptsize
  \setlength{\figurewidth}{.15\textwidth}
  \setlength{\figureheight}{\figurewidth}
  \begin{tikzpicture}

    \tikzstyle{fig} = [draw=white,minimum size=\figurewidth,inner sep=0pt]

    \newcommand{\figg}[1]{\includegraphics[width=.97\figurewidth]{./supplement/feature_manipulation/smooth/#1.jpg}};
  
    \foreach \i in {5276,...,5281} {

      \node[fig] at ({\figurewidth*mod((\i - 5276),6)},{-\figureheight*0}) {\figg{male/\i}};
    }
    \foreach \i in {5540,...,5545} {

      \node[fig] at ({\figurewidth*mod((\i - 5540),6)},{-\figureheight*1}) {\figg{bald/\i}};
    }
    \foreach \i in {5204,...,5209} {

      \node[fig] at ({\figurewidth*mod((\i - 5204),6)},{-\figureheight*2}) {\figg{smile/\i}};
    }
    \foreach \i in {5684,...,5689} {

      \node[fig] at ({\figurewidth*mod((\i - 5684),6)},{-\figureheight*3}) {\figg{sunglasses/\i}};
    }

    \foreach \x [count=\i] in {Input,Reconstruction,Gradual change $\rightarrow$} {  
      \node[fill=white,inner sep=0, minimum width=4mm] at ({\figurewidth*(\i-1)},-3.6\figureheight) {\x};
    }

    \foreach \x [count=\i] in {Female $\rightarrow$ Male,Make bald,Add smile,Add sunglasses} {  
      \node[fill=white,inner sep=0, minimum width=4mm, rotate=90] at (-0.6\figurewidth,{-\figureheight*\i+\figureheight}) {\x};
    }
  \end{tikzpicture}
  \caption{Balanced \PIONEER gradual feature manipulation (\CelebAHQ) at $256{\times}256$ resolution by increasing $\lambda$ for a single feature. Column~1: Input; Column~2: Reconstruction ($\lambda=0$); Columns~3--6: $\lambda$ increasing. Row~1: Female $\rightarrow$ Male; Row~2: Make bald; Row~3: Add smile; Row~4: Add sunglasses.}
  \label{fig:manipulate2}
\end{figure*}
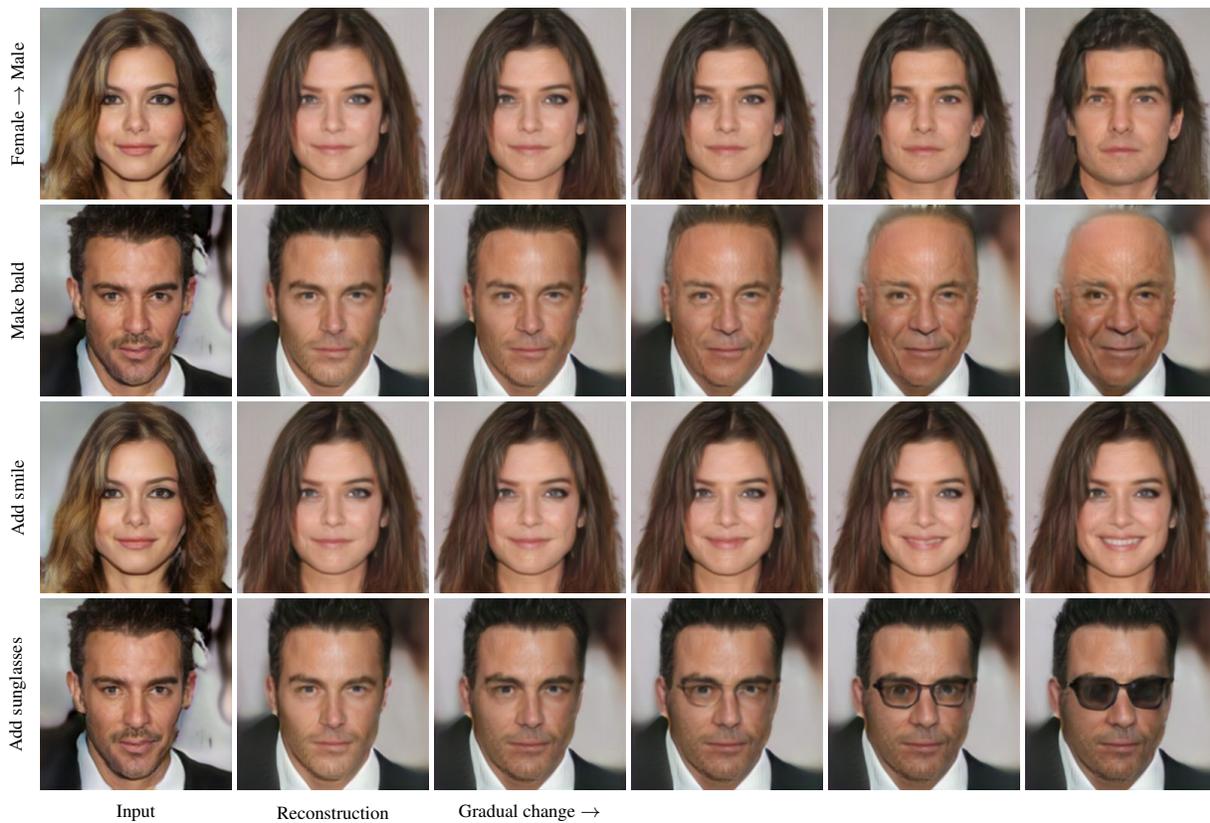

\begin{figure*}[h]
  \centering\scriptsize
  \setlength{\figurewidth}{.15\textwidth}
  \setlength{\figureheight}{\figurewidth}
  \begin{tikzpicture}

    \tikzstyle{fig} = [draw=white,minimum size=\figurewidth,inner sep=0pt]

    \newcommand{\figg}[1]{\includegraphics[width=.97\figurewidth]{./supplement/feature_manipulation/discrete/#1.jpg}};

    \node[fig] at ({\figurewidth*0},{-\figureheight*0}) {\figg{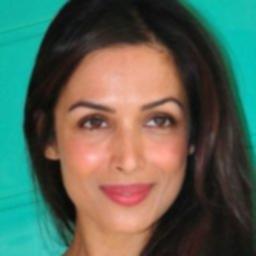}};
    \node[fig] at ({\figurewidth*1},{-\figureheight*0}) {\figg{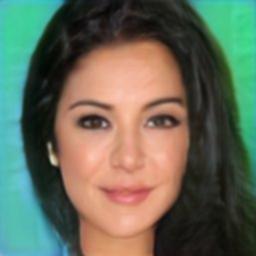}};
    \node[fig] at ({\figurewidth*2},{-\figureheight*0}) {\figg{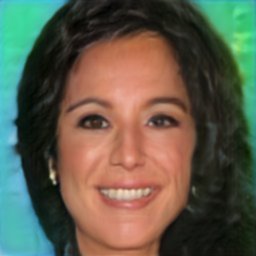}};
    \node[fig] at ({\figurewidth*3},{-\figureheight*0}) {\figg{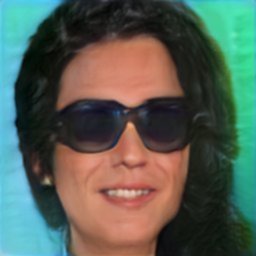}};
    \node[fig] at ({\figurewidth*4},{-\figureheight*0}) {\figg{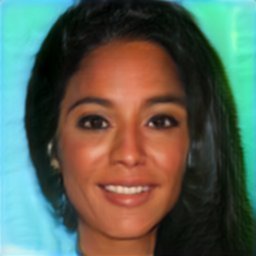}};
    \node[fig] at ({\figurewidth*5},{-\figureheight*0}) {\figg{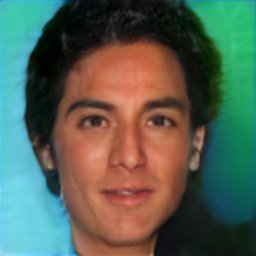}};

    \node[fig] at ({\figurewidth*0},{-\figureheight*1}) {\figg{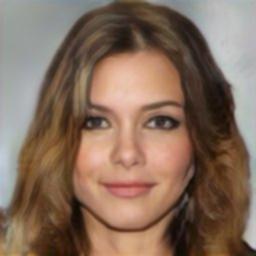}};
    \node[fig] at ({\figurewidth*1},{-\figureheight*1}) {\figg{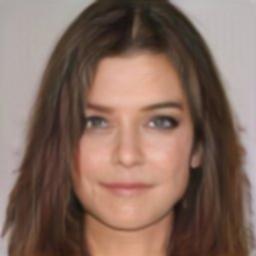}};
    \node[fig] at ({\figurewidth*2},{-\figureheight*1}) {\figg{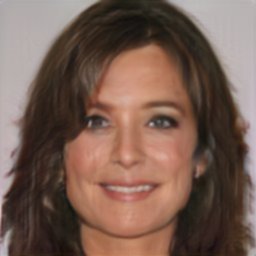}};
    \node[fig] at ({\figurewidth*3},{-\figureheight*1}) {\figg{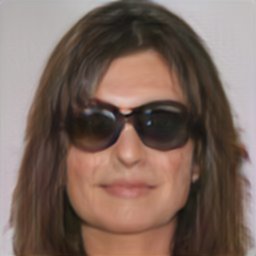}};
    \node[fig] at ({\figurewidth*4},{-\figureheight*1}) {\figg{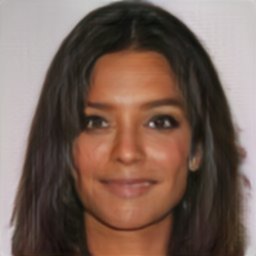}};
    \node[fig] at ({\figurewidth*5},{-\figureheight*1}) {\figg{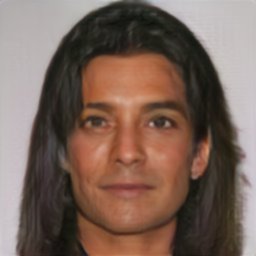}};

    \node[fig] at ({\figurewidth*0},{-\figureheight*2}) {\figg{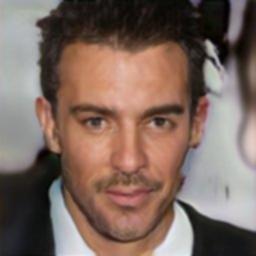}};
    \node[fig] at ({\figurewidth*1},{-\figureheight*2}) {\figg{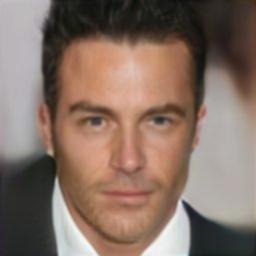}};
    \node[fig] at ({\figurewidth*2},{-\figureheight*2}) {\figg{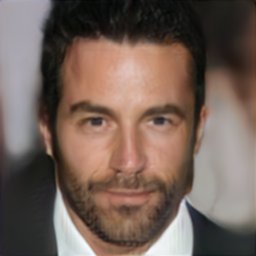}};
    \node[fig] at ({\figurewidth*3},{-\figureheight*2}) {\figg{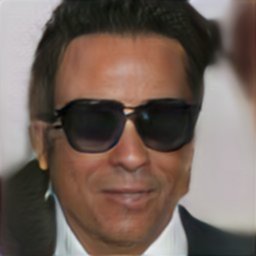}};
    \node[fig] at ({\figurewidth*4},{-\figureheight*2}) {\figg{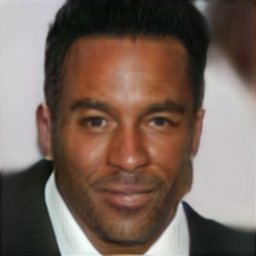}};
    \node[fig] at ({\figurewidth*5},{-\figureheight*2}) {\figg{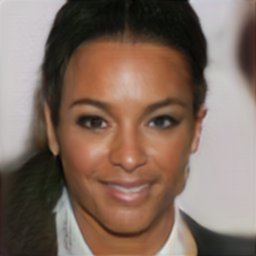}};
    
    \node[fig] at ({\figurewidth*0},{-\figureheight*3}) {\figg{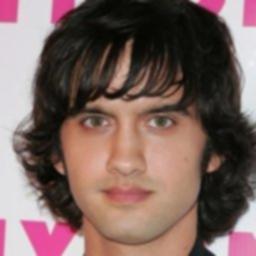}};
    \node[fig] at ({\figurewidth*1},{-\figureheight*3}) {\figg{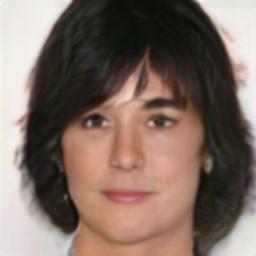}};
    \node[fig] at ({\figurewidth*2},{-\figureheight*3}) {\figg{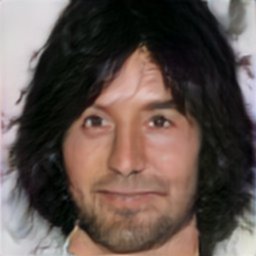}};
    \node[fig] at ({\figurewidth*3},{-\figureheight*3}) {\figg{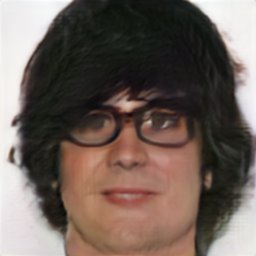}};
    \node[fig] at ({\figurewidth*4},{-\figureheight*3}) {\figg{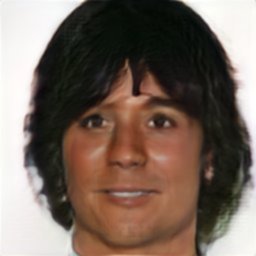}};
    \node[fig] at ({\figurewidth*5},{-\figureheight*3}) {\figg{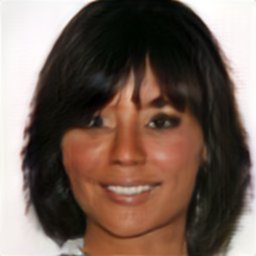}};

    \node[fig] at ({\figurewidth*0},{-\figureheight*4}) {\figg{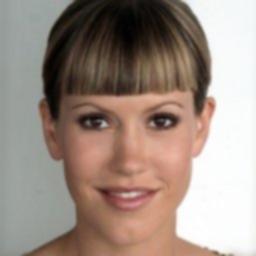}};
    \node[fig] at ({\figurewidth*1},{-\figureheight*4}) {\figg{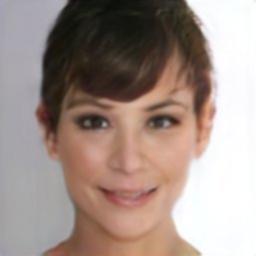}};
    \node[fig] at ({\figurewidth*2},{-\figureheight*4}) {\figg{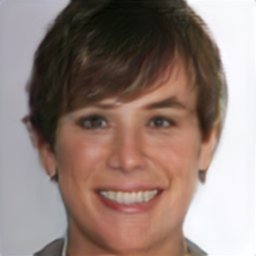}};
    \node[fig] at ({\figurewidth*3},{-\figureheight*4}) {\figg{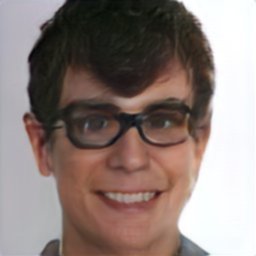}};
    \node[fig] at ({\figurewidth*4},{-\figureheight*4}) {\figg{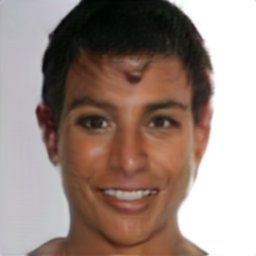}};
    \node[fig] at ({\figurewidth*5},{-\figureheight*4}) {\figg{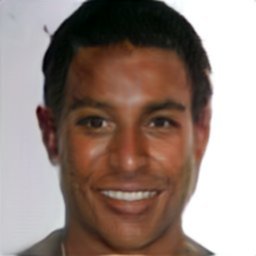}};
    
    \node[fig] at ({\figurewidth*0},{-\figureheight*5}) {\figg{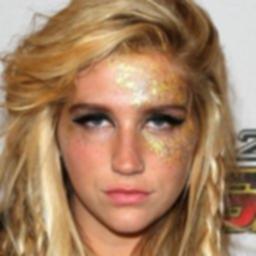}};
    \node[fig] at ({\figurewidth*1},{-\figureheight*5}) {\figg{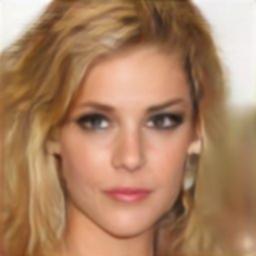}};
    \node[fig] at ({\figurewidth*2},{-\figureheight*5}) {\figg{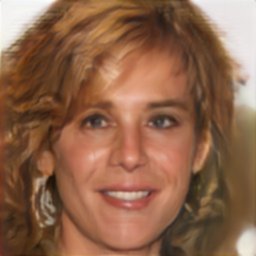}};
    \node[fig] at ({\figurewidth*3},{-\figureheight*5}) {\figg{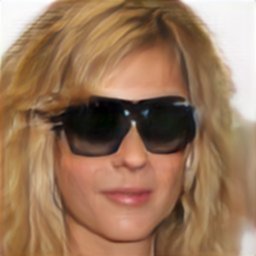}};
    \node[fig] at ({\figurewidth*4},{-\figureheight*5}) {\figg{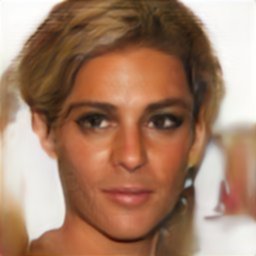}};
    \node[fig] at ({\figurewidth*5},{-\figureheight*5}) {\figg{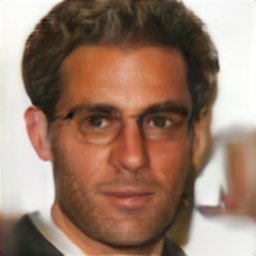}};
    
    \node[fig] at ({\figurewidth*0},{-\figureheight*6}) {\figg{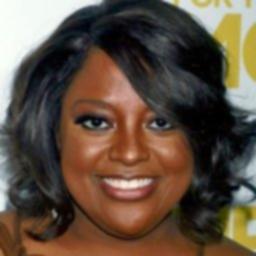}};
    \node[fig] at ({\figurewidth*1},{-\figureheight*6}) {\figg{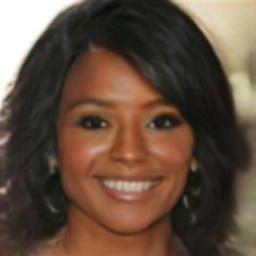}};
    \node[fig] at ({\figurewidth*2},{-\figureheight*6}) {\figg{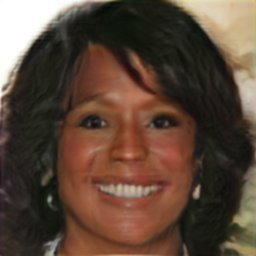}};
    \node[fig] at ({\figurewidth*3},{-\figureheight*6}) {\figg{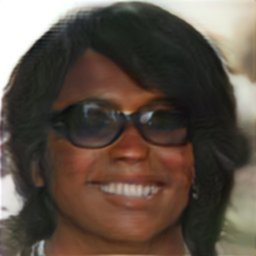}};
    \node[fig] at ({\figurewidth*4},{-\figureheight*6}) {\figg{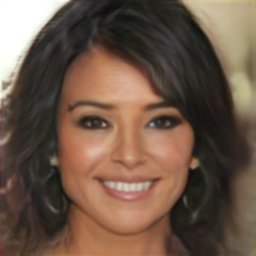}};
    \node[fig] at ({\figurewidth*5},{-\figureheight*6}) {\figg{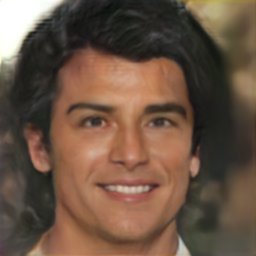}};

    \foreach \x [count=\i] in {Input,Reconstruction,Older/Add beard,Older+Glasses,Darken/whiten,Switch sex+Darken/whiten} {  
      \node[fill=white,inner sep=0, minimum width=4mm] at ({\figurewidth*(\i-1)},-6.6\figureheight) {\x};
    }    
  \end{tikzpicture}
  \caption{Balanced \PIONEER discrete feature manipulation (\CelebAHQ) at $256{\times}256$ resolution by adding various feature vectors. Column~1: Input; Column~2: Reconstruction ($\lambda=0$); Column~3: Make older (females) or Add beard (males); Colum~4: Make older + Add (sun)glasses; Column~5: Darken/whiten the skin; Column~6: Switch sex + Darken/whiten the skin.}
  \label{fig:manipulate3}
\end{figure*}

\noindent
Following the method in Sec.~\ref{sec:disentanglement}, we take the same \CelebAHQ model that was trained in a completely unsupervised manner, and apply the feature vectors from the latent space as in Fig.~\ref{fig:manipulate}, but this time showing how each feature transforms the (reconstruction of the) input image gradually as a function of $\lambda$ (Fig.~\ref{fig:manipulate2}). We also provide more examples of the features in Fig.~\ref{fig:manipulate} applied to other images, and other features computed with the same method, and their combinations (Fig.~\ref{fig:manipulate3}).

\section{Latent space interpolations}
\noindent
We show 4-way interpolation examples for uncurated \CelebAHQ test set images, following the same method as used for Fig.~\ref{fig:hex}, but with evenly spaced (spherical) interpolation between the reconstructions of each of the input images in the corners (Fig.~\ref{fig:interpolation2}--\ref{fig:interpolation4}). Fig.~\ref{fig:interpolation3} represents a failure case.

\begin{figure*}[!t]
        \centering

        \resizebox{\textwidth}{!}{%
        \begin{tikzpicture}[inner sep=0]

          \newcommand{\figg}[2]{\includegraphics[width=1cm]{fig/fig5/interpolations_6_24520000_1_#1x#2.jpg}}

          \foreach \i in {0,...,7} {
            \foreach \j in {0,...,7} {
              \node (\i-\j) [] at (\j,-\i) {\figg{\i}{\j}};
            }
          }

          \newcommand{\insquare}[3]{\node [minimum width=1.5cm,minimum height=1.5cm, rounded corners=3pt,path picture={\node at (path picture bounding box.center){\includegraphics[width=1.5cm]{#3}};}] at (#1,#2) {};};

          \insquare{-1.5}{-0.25}{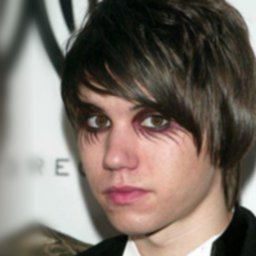}
          \insquare{8.50}{-0.25}{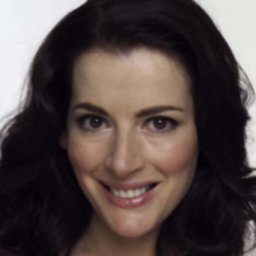}
          \insquare{-1.5}{-6.75}{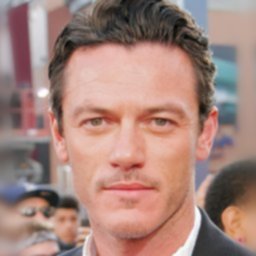}
          \insquare{8.50}{-6.75}{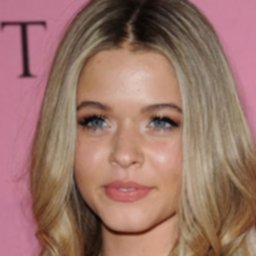}
          
        \end{tikzpicture}}
        \caption{Interpolation between random uncurated test set \CelebAHQ images in $256{\times }256$. The model captures most of the salient features, although fails with some details such as the unsual make-up of the top-left person. Note that the moderate rotation angles of the faces are almost perfectly preserved, and the intermediate faces are rotated to the correct degree.}
        \label{fig:interpolation}
\end{figure*}

\begin{figure*}[!t]
  \centering

  \resizebox{\textwidth}{!}{%
  \begin{tikzpicture}[inner sep=0]

    \newcommand{\figg}[2]{\includegraphics[width=1cm]{./supplement/interpolations/interpolations_6_25480001_1_#1x#2.jpg}}

    \foreach \i in {0,...,7} {
      \foreach \j in {0,...,7} {
        \node (\i-\j) [] at (\j,-\i) {\figg{\i}{\j}};
      }
    }

    \newcommand{\insquare}[3]{\node [minimum width=1.5cm,minimum height=1.5cm, rounded corners=3pt,path picture={\node at (path picture bounding box.center){\includegraphics[width=1.5cm]{#3}};}] at (#1,#2) {};};

    \insquare{-1.5}{-0.25}{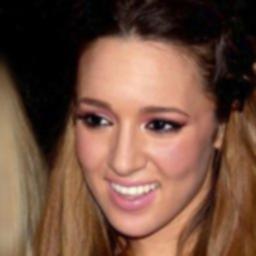}
    \insquare{8.50}{-0.25}{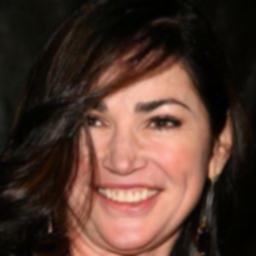}
    \insquare{-1.5}{-6.75}{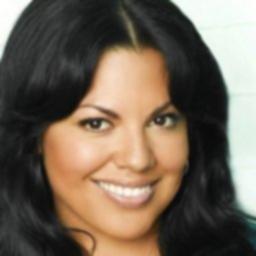}
    \insquare{8.50}{-6.75}{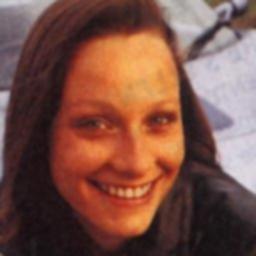}
    
  \end{tikzpicture}}
  \caption{Interpolation between random uncurated test set \CelebAHQ images in $256{\times }256$.}
  \label{fig:interpolation2}
\end{figure*}

\begin{figure*}[!t]
  \centering

  \resizebox{\textwidth}{!}{%
  \begin{tikzpicture}[inner sep=0]

    \newcommand{\figg}[2]{\includegraphics[width=1cm]{./supplement/interpolations/interpolations_6_25480002_1_#1x#2.jpg}}

    \foreach \i in {0,...,7} {
      \foreach \j in {0,...,7} {
        \node (\i-\j) [] at (\j,-\i) {\figg{\i}{\j}};
      }
    }

    \newcommand{\insquare}[3]{\node [minimum width=1.5cm,minimum height=1.5cm, rounded corners=3pt,path picture={\node at (path picture bounding box.center){\includegraphics[width=1.5cm]{#3}};}] at (#1,#2) {};};

    \insquare{-1.5}{-0.25}{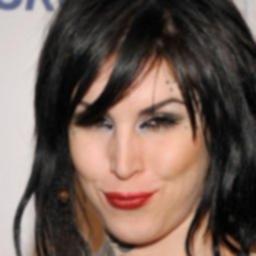}
    \insquare{8.50}{-0.25}{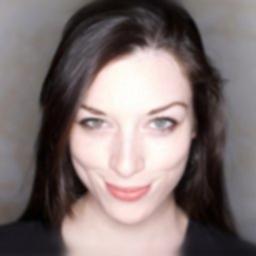}
    \insquare{-1.5}{-6.75}{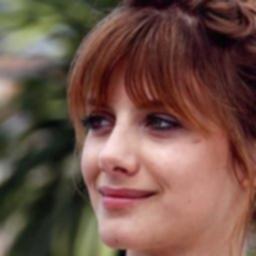}
    \insquare{8.50}{-6.75}{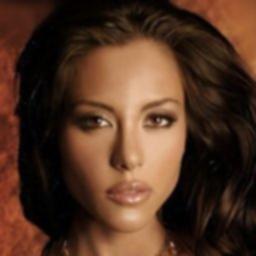}
    
  \end{tikzpicture}}
  \caption{Interpolation between random uncurated test set \CelebAHQ images in $256{\times }256$. Bottom-left image reconstruction is clearly inadequate.}
  \label{fig:interpolation3}
\end{figure*}

\begin{figure*}[!t]
  \centering

  \resizebox{\textwidth}{!}{%
  \begin{tikzpicture}[inner sep=0]

    \newcommand{\figg}[2]{\includegraphics[width=1cm]{./supplement/interpolations/interpolations_6_25480003_1_#1x#2.jpg}}

    \foreach \i in {0,...,7} {
      \foreach \j in {0,...,7} {
        \node (\i-\j) [] at (\j,-\i) {\figg{\i}{\j}};
      }
    }

    \newcommand{\insquare}[3]{\node [minimum width=1.5cm,minimum height=1.5cm, rounded corners=3pt,path picture={\node at (path picture bounding box.center){\includegraphics[width=1.5cm]{#3}};}] at (#1,#2) {};};

    \insquare{-1.5}{-0.25}{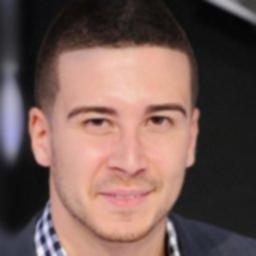}
    \insquare{8.50}{-0.25}{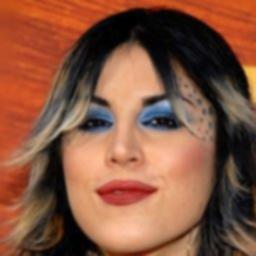}
    \insquare{-1.5}{-6.75}{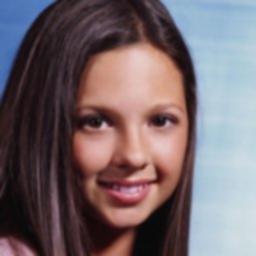}
    \insquare{8.50}{-6.75}{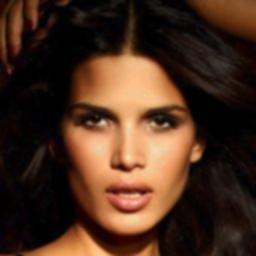}
    
  \end{tikzpicture}}
  \caption{Interpolation between random uncurated test set \CelebAHQ images in $256{\times }256$.}
  \label{fig:interpolation4}
\end{figure*}

\section{Random samples} 
\begin{figure*}[h]
            \centering\scriptsize
            \setlength{\figurewidth}{.145\textwidth}
            \setlength{\figureheight}{\figurewidth}
            \begin{tikzpicture}

              \tikzstyle{fig} = [draw=white,minimum size=\figurewidth,inner sep=0pt]

              \newcommand{\figg}[1]{\includegraphics[width=.97\figurewidth]{./supplement/celebaHQ_samples/pine/25480001_#1.jpg}};
            
              \foreach \i in {9539,...,9558} {

                \node[fig] at ({\figurewidth*int((\i - 9539)/4)},{-\figureheight*mod((\i - 9539),4)}) {\figg{\i}};
              }
            \end{tikzpicture}
            \caption{Balanced \PIONEER (ours) random samples (\CelebAHQ) at $256{\times}256$ resolution.}
            \label{fig:genPINE2celeba}
\end{figure*}
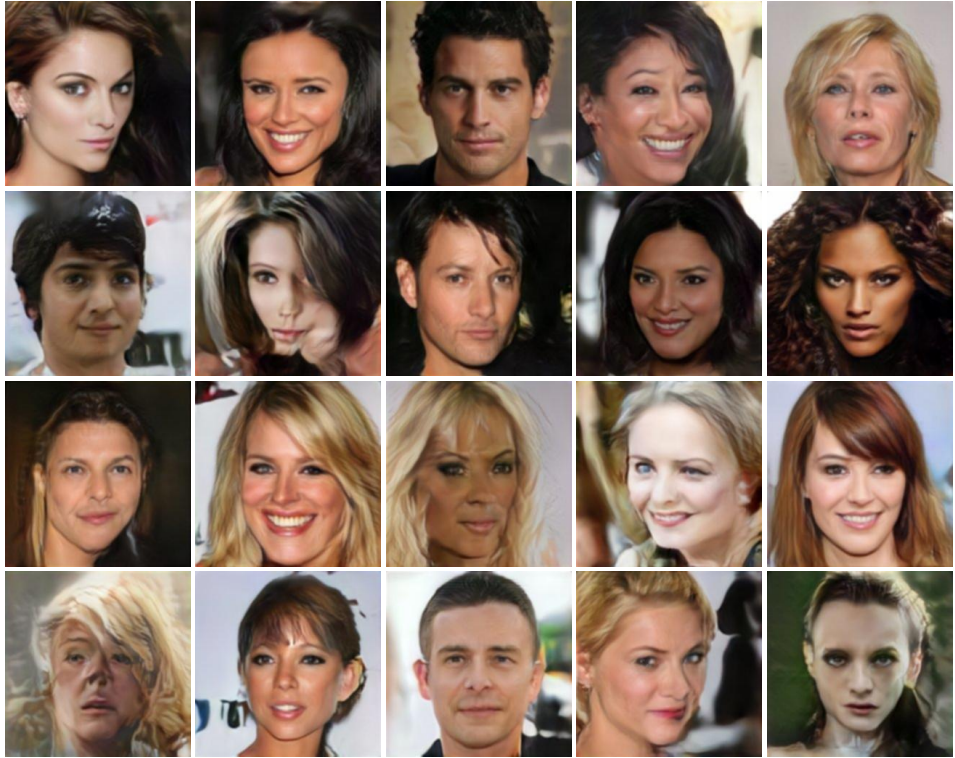

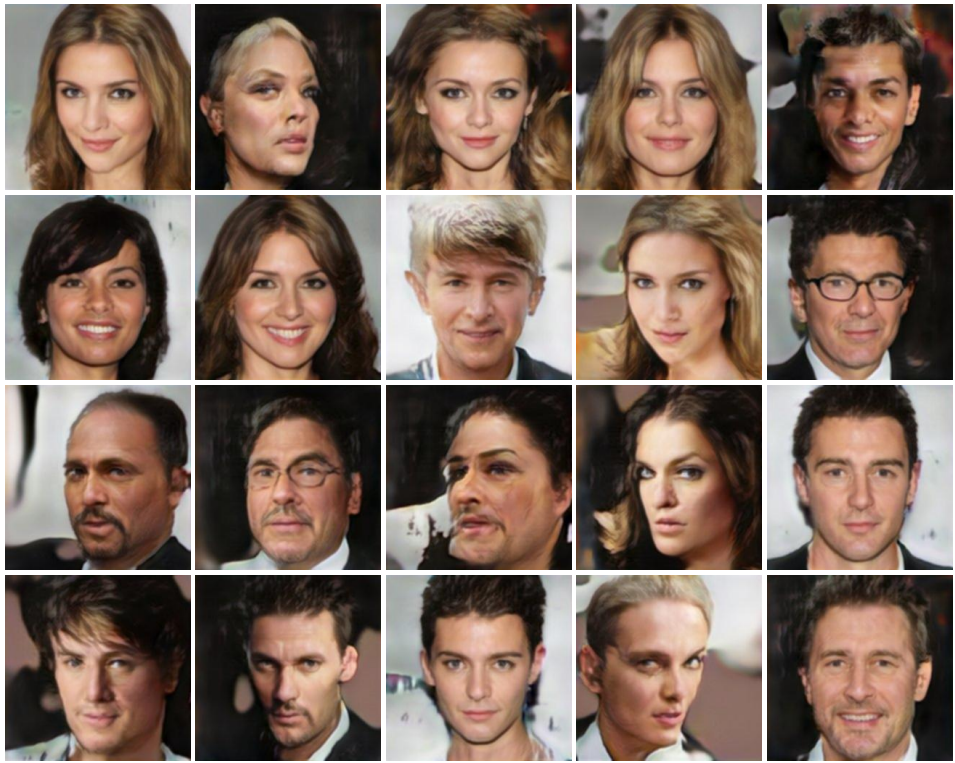
\begin{figure*}[h]
  \centering\scriptsize
  \setlength{\figurewidth}{.145\textwidth}
  \setlength{\figureheight}{\figurewidth}
  \begin{tikzpicture}

    \tikzstyle{fig} = [draw=white,minimum size=\figurewidth,inner sep=0pt]

    \newcommand{\figg}[1]{\includegraphics[width=.97\figurewidth]{./supplement/celebaHQ_samples/pine_old/27480001_#1.jpg}};
  
    \foreach \i in {0,...,19} {

    \node[fig] at ({\figurewidth*int((\i)/4)},{-\figureheight*mod((\i),4)}) {\figg{\i}};
    }
  \end{tikzpicture}
  \caption{Baseline \PIONEER random samples (\CelebAHQ) at $256{\times}256$ resolution.}
  \label{fig:genPINEceleba}
\end{figure*}

\begin{figure*}[h]
  \centering\scriptsize
  \setlength{\figurewidth}{.145\textwidth}
  \setlength{\figureheight}{\figurewidth}
  \begin{tikzpicture}

    \tikzstyle{fig} = [draw=white,minimum size=\figurewidth,inner sep=0pt]

    \newcommand{\figg}[1]{\includegraphics[width=.97\figurewidth]{./supplement/celebaHQ_samples/glow_07/rand0#1.jpg}};
  
    \foreach \i in {719,...,738} {

      \node[fig] at ({\figurewidth*int((\i - 719)/4)},{-\figureheight*mod((\i - 719),4)}) {\figg{\i}};
    }
  \end{tikzpicture}
  \caption{GLOW random samples (\CelebAHQ) at $256{\times}256$ resolution, temperature $T=0.7$.}
  \label{fig:genGLOWceleba}
\end{figure*}
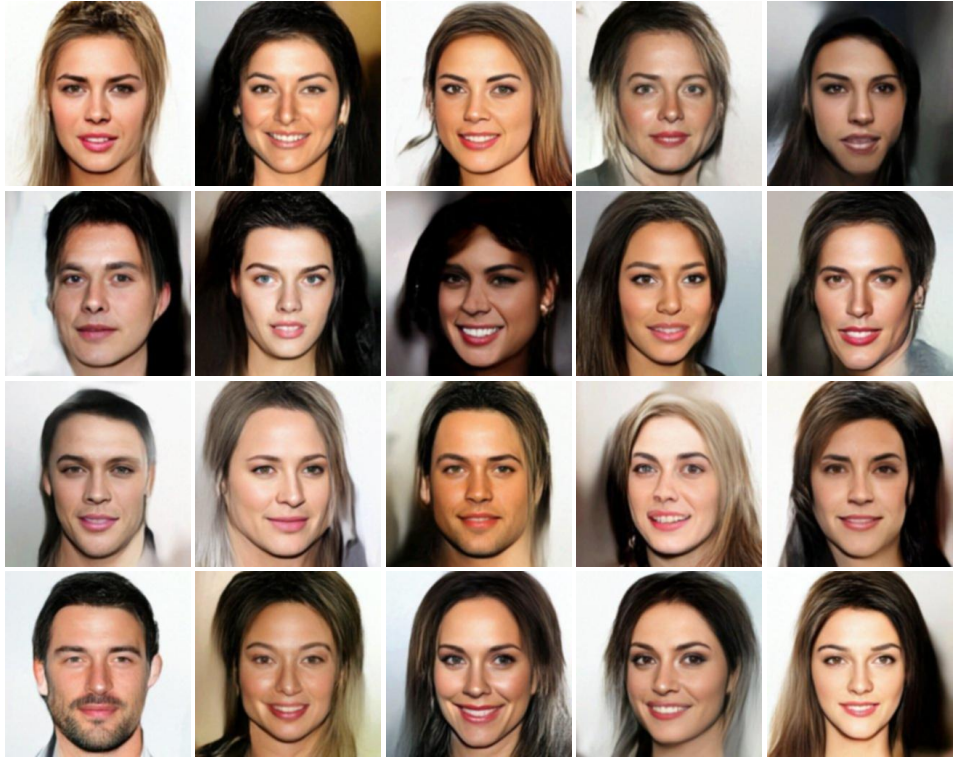

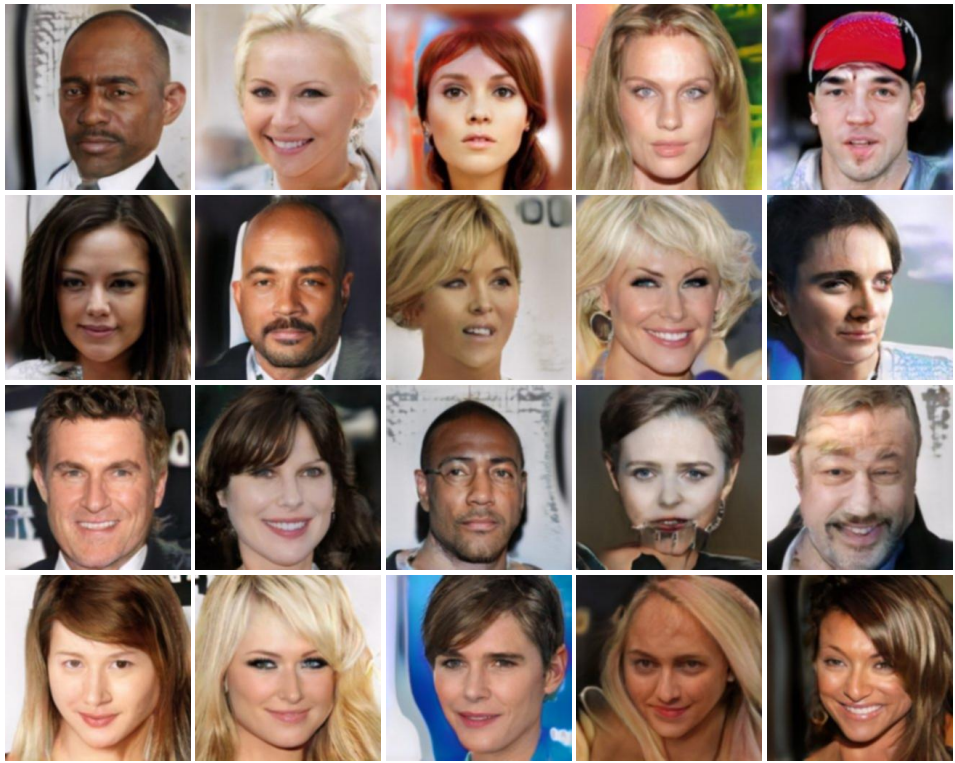
\begin{figure*}[h]
  \centering\scriptsize
  \setlength{\figurewidth}{.145\textwidth}
  \setlength{\figureheight}{\figurewidth}
  \begin{tikzpicture}

    \tikzstyle{fig} = [draw=white,minimum size=\figurewidth,inner sep=0pt]

    \newcommand{\figg}[1]{\includegraphics[width=.97\figurewidth]{./supplement/celebaHQ_samples/pggan/015-pgan-celeba-preset-v2-4gpus-fp32-network-snapshot-010000-0000#1.jpg}};
  
    \foreach \i in {10,...,29} {

      \node[fig] at ({\figurewidth*int((\i - 10)/4)},{-\figureheight*mod((\i - 10),4)}) {\figg{\i}};
    }
  \end{tikzpicture}
  \caption{Progressively Growing GAN (PGGAN) random samples (\CelebAHQ) at $256{\times}256$ resolution.}
  \label{fig:genPGGANceleba}
\end{figure*}

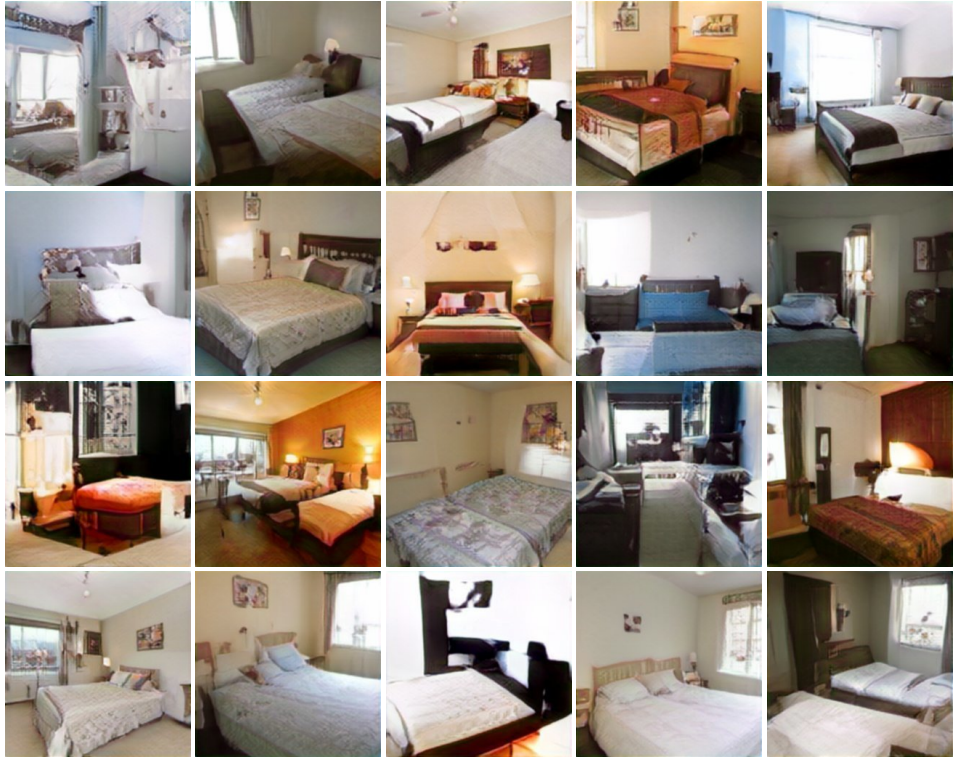
\begin{figure*}[h]
  \centering\scriptsize
  \setlength{\figurewidth}{.145\textwidth}
  \setlength{\figureheight}{\figurewidth}
  \begin{tikzpicture}

    \tikzstyle{fig} = [draw=white,minimum size=\figurewidth,inner sep=0pt]

    \newcommand{\figg}[1]{\includegraphics[width=.97\figurewidth]{./supplement/lsun_samples/balanced_pioneer/final/25400001_#1.jpg}};
  
    \foreach \i in {7287,...,7306} {

      \node[fig] at ({\figurewidth*int((\i - 7287)/4)},{-\figureheight*mod((\i - 7287),4)}) {\figg{\i}};
    }
  \end{tikzpicture}
  \caption{Balanced \PIONEER (ours) random samples (\LSUN Bedrooms) at $256{\times}256$ resolution.}
  \label{fig:genPINE2lsun}
\end{figure*}

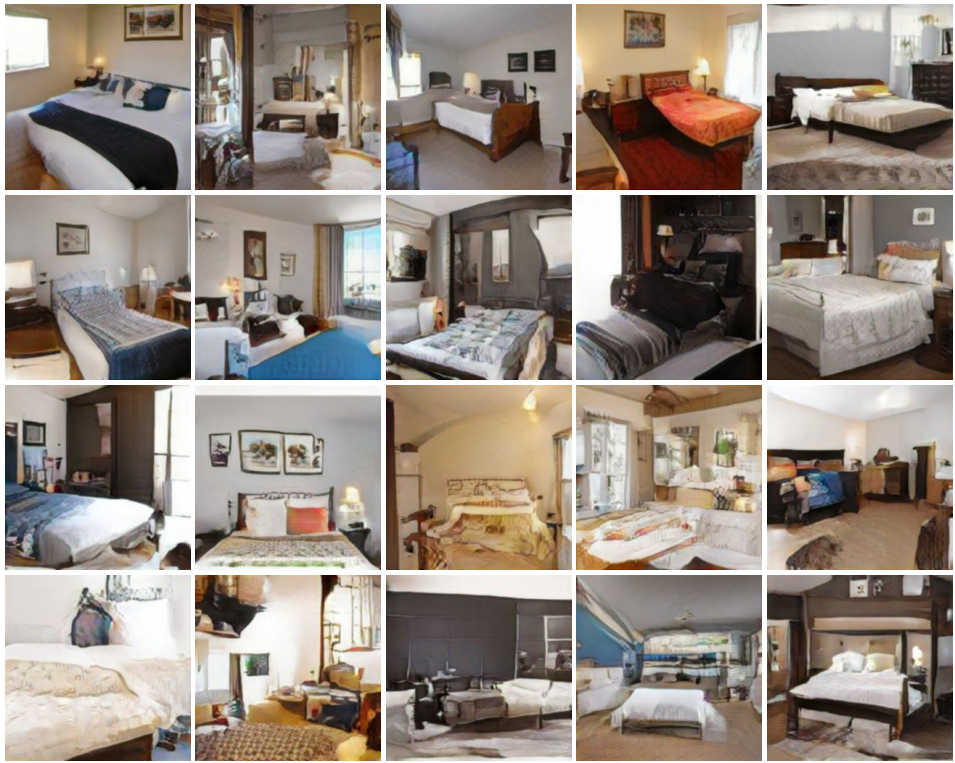
\begin{figure*}[h]
\centering\scriptsize
\setlength{\figurewidth}{.145\textwidth}
\setlength{\figureheight}{\figurewidth}
\begin{tikzpicture}
\tikzstyle{fig} = [draw=white,minimum size=\figurewidth,inner sep=0pt]

\newcommand{\figg}[1]{\includegraphics[width=.97\figurewidth]{./supplement/lsun_samples/baseline/B27384001_#1.jpg}};

\foreach \i in {9551,...,9570} {
\node[fig] at ({\figurewidth*int((\i - 9551)/4)},{-\figureheight*mod((\i - 9551),4)}) {\figg{\i}};
}
\end{tikzpicture}
\caption{Baseline \PIONEER random samples (\LSUN Bedrooms) at $256{\times}256$ resolution.}
\label{fig:genPINElsun}
\end{figure*}

\noindent
We show more random generated samples for \CelebAHQ using Balanced \PIONEER (ours) (Fig.~\ref{fig:genPINE2celeba}), the baseline \PIONEER (Fig.~\ref{fig:genPINEceleba}), GLOW (Fig.~\ref{fig:genGLOWceleba}) and PGGAN (Fig.~\ref{fig:genPGGANceleba}). Similarly, we show more random samples for \LSUN using Balanced \PIONEER (ours) (Fig.~\ref{fig:genPINE2lsun}) and the baseline \PIONEER (Fig.~\ref{fig:genPINElsun}).

\section{Reconstructions} 
\noindent
We show more uncurated examples of \CelebAHQ reconstructions, comparing Balanced \PIONEER (ours) against the baseline \PIONEER (Fig.~\ref{fig:reconstructions2}--\ref{fig:reconstructions3}).

\begin{figure*}[!t]
  \centering\footnotesize
  \setlength{\figurewidth}{.15\textwidth}
  \setlength{\figureheight}{\figurewidth}

  \begin{tikzpicture}[inner sep=0]

  \tikzstyle{arrow} = [draw=black!10, single arrow, minimum height=10mm, minimum width=3mm, single arrow head extend=1mm, fill=black!10, anchor=center, rotate=-90, inner sep=2pt]

  \foreach \x [count=\i] in {9781,9782,9787,9786,9784,9785} {
     \node[] at ({\figurewidth*\i},{0*\figureheight}) {\includegraphics[width=\figurewidth]{./supplement/reconstructions/\x_orig.jpg}};
     \node[] at ({\figurewidth*\i},{-1.2\figureheight}) {\includegraphics[width=\figurewidth]{./supplement/reconstructions/\x_pine.jpg}};
     \node[] at ({\figurewidth*\i},{-2.2\figureheight}) {\includegraphics[width=\figurewidth]{./supplement/reconstructions/\x_pine_old.jpg}};
     \node[arrow] at ({\figurewidth*\i},{-0.6\figureheight}) {};}

  \end{tikzpicture}
  \caption{More examples of reconstruction quality in $256{\times}256$ resolution with typical images from the \CelebAHQ test set (top row), by our balanced \PIONEER (middle) and baseline \PIONEER (bottom). Here, the input images are encoded into 512-dimensional latent feature vector and decoded back to the original dimensionality (middle and bottom rows). The encoding--decoding of balanced \PIONEER tends to preserve facial features, orientation, expressions, and hair style. Small mistakes can still be observed, especially in male subjects.\protect\\[.5em]}
  \label{fig:reconstructions2}
\end{figure*}

\begin{figure*}[!t]
  \centering\footnotesize
  \setlength{\figurewidth}{.15\textwidth}
  \setlength{\figureheight}{\figurewidth}

  \begin{tikzpicture}[inner sep=0]

  \tikzstyle{arrow} = [draw=black!10, single arrow, minimum height=10mm, minimum width=3mm, single arrow head extend=1mm, fill=black!10, anchor=center, rotate=-90, inner sep=2pt]

  \foreach \x [count=\i] in {9791,9800,9801,9802,9803,9804} {
     \node[] at ({\figurewidth*\i},{0*\figureheight}) {\includegraphics[width=\figurewidth]{./supplement/reconstructions/\x_orig.jpg}};
     \node[] at ({\figurewidth*\i},{-1.2\figureheight}) {\includegraphics[width=\figurewidth]{./supplement/reconstructions/\x_pine.jpg}};
     \node[] at ({\figurewidth*\i},{-2.2\figureheight}) {\includegraphics[width=\figurewidth]{./supplement/reconstructions/\x_pine_old.jpg}};
     \node[arrow] at ({\figurewidth*\i},{-0.6\figureheight}) {};}

  \end{tikzpicture}
  \caption{More examples of reconstruction quality in $256{\times}256$ resolution with typical images from the \CelebAHQ test set (top row), by our balanced \PIONEER (middle) and baseline \PIONEER (bottom). Here, the input images are encoded into 512-dimensional latent feature vector and decoded back to the original dimensionality (middle and bottom rows). The encoding--decoding of balanced \PIONEER tends to preserve facial features, orientation, expressions, and hair style. Small mistakes can still be observed, especially in male subjects.\protect\\[.5em]}
  \label{fig:reconstructions3}
\end{figure*}

\section{Feature transformation videos}
\begin{figure*}[h]
  \centering\scriptsize
  \setlength{\figurewidth}{.15\textwidth}
  \setlength{\figureheight}{\figurewidth}
  \begin{tikzpicture}

    \tikzstyle{fig} = [draw=white,minimum size=\figurewidth,inner sep=0pt]

    \newcommand{\figg}[1]{\includegraphics[width=.97\figurewidth]{./supplement/feature_manipulation/discrete/#1.jpg}};

    \node[fig] at ({\figurewidth},{-\figureheight*0}) {\figg{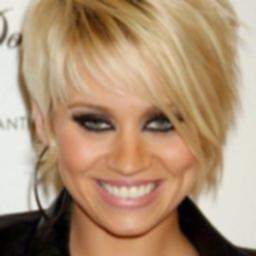}};
    \node[fig] at ({\figurewidth*mod(2,6)},{-\figureheight*0}) {\figg{10160}};
    \node[fig] at ({\figurewidth*mod(3,6)},{-\figureheight*0}) {\figg{10155}};
    \node[fig] at ({\figurewidth*mod(4,6)},{-\figureheight*0}) {\figg{10150}};

  \end{tikzpicture}
  \caption{\CelebAHQ test set images used as input for the image transformation videos.}
  \label{fig:videoinputs}
\end{figure*}

\noindent 
The attached video (see \url{https://aaltovision.github.io/balanced-pioneer}) demonstrates various gradual feature transformations (as in Fig.~\ref{fig:manipulate2}). Each transformation showcases $\lambda$ varying on a subrange of $[-2.0, 2.0]$, applied on the original test set images shown in Fig.~\ref{fig:videoinputs}.

\end{document}